\documentclass[10pt,twocolumn,letterpaper]{article}

\usepackage{iccv}              

\definecolor{iccvblue}{rgb}{0.21,0.49,0.74}
\usepackage[pagebackref,breaklinks,colorlinks,allcolors=iccvblue]{hyperref}
\usepackage{array}

\newcolumntype{L}[1]{>{\raggedright\let\newline\\\arraybackslash\hspace{0pt}}m{#1}}
\newcolumntype{C}[1]{>{\centering\let\newline\\\arraybackslash\hspace{0pt}}m{#1}}
\newcolumntype{R}[1]{>{\raggedleft\let\newline\\\arraybackslash\hspace{0pt}}m{#1}}

\def \E {\mathbb{E}}

\def \path {\mathit{path}}

\newcommand{\abs}[1]{\left\vert#1\right\vert}

\usepackage{amsmath,amsfonts,bm,amssymb}

\definecolor{uggblue}{RGB}{96, 150, 179}
\definecolor{uggred}{RGB}{223, 120, 97}
\definecolor{ugggreen}{RGB}{148, 180, 159}

\usepackage{pifont}
\newcommand{\cmark}{\ding{51}}%
\newcommand{\xmark}{\ding{55}}%

\usepackage{multirow}
\usepackage[accsupp]{axessibility}

\newcommand{\name}{HUMOTO\xspace}

\newcommand{\namelongunderlined}{\underline{Hu}man \underline{Mot}ions with \underline{O}bjects\xspace}
\def\paperID{1782} 
\def\confName{ICCV}
\def\confYear{2025}

\usepackage{lipsum}

\newcommand\blfootnote[1]{%
  \begingroup
  \renewcommand\thefootnote{}\footnote{#1}%
  \addtocounter{footnote}{-1}%
  \endgroup
}

\title{HUMOTO: A 4D Dataset of Mocap Human Object Interactions}

\author{Jiaxin Lu$^{1,2}$\thanks{}, Chun-Hao Paul Huang$^2$, Uttaran Bhattacharya$^2$, Qixing Huang$^1$, Yi Zhou$^2$\\
$^1$University of Texas at Austin, $^2$ Adobe Research\\
{\tt\small lujiaxin@utexas.edu, \{chunhaoh, ubhattac, yizho\}@adobe.com, huangqx@cs.utexas.edu}
}

\begin{document}

\twocolumn[{%
\renewcommand\twocolumn[1][]{#1}%
\maketitle
\begin{center}
    \centering
    \captionsetup{type=figure}
    \includegraphics[width=\linewidth]{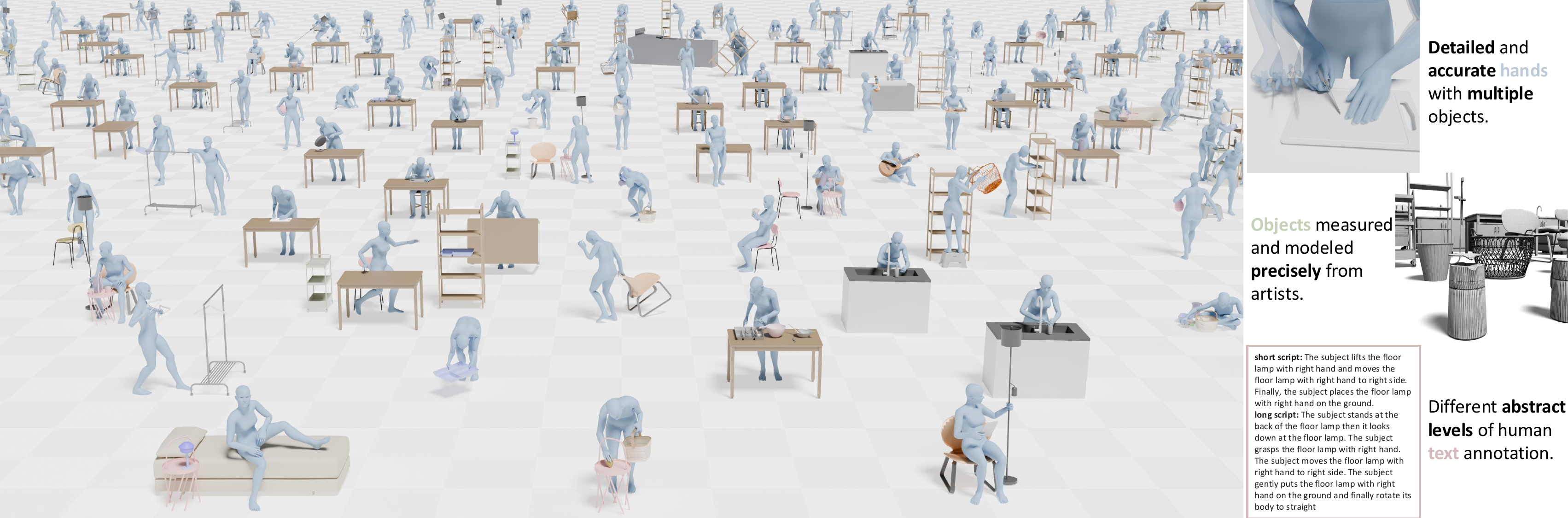}
    \captionof{figure}{\textbf{Overview of the \name dataset.} The dataset contains mocap 4D human-object interaction animations with multiple objects. The unique features of the dataset include its detailed, accurate interaction modeling, specifically the detailed hand pose. The objects are precisely modeled by artists. We additionally provide different abstract levels of text annotation for the interactions.}
    \label{fig:overview}
\end{center}%
}]
\blfootnote{$^*$ The work was mainly conducted at Adobe Research.}

\begin{abstract}
We present \namelongunderlined (\name), a high-fidelity dataset of human-object interactions for motion generation, computer vision, and robotics applications. Featuring 735 sequences (7,875 seconds at 30 fps), \name captures interactions with 63 precisely modeled objects and 72 articulated parts. Our innovations include a scene-driven LLM scripting pipeline creating complete, purposeful tasks with natural progression, and a mocap-and-camera recording setup to effectively handle occlusions. Spanning diverse activities from cooking to outdoor picnics, \name preserves both physical accuracy and logical task flow. Professional artists rigorously clean and verify each sequence, minimizing foot sliding and object penetrations. We also provide benchmarks compared to other datasets. \name's comprehensive full-body motion and simultaneous multi-object interactions address key data-capturing challenges and provide opportunities to advance realistic human-object interaction modeling across research domains with practical applications in animation, robotics, and embodied AI systems. 
Project Page: \url{https://jiaxin-lu.github.io/humoto/}.
\end{abstract}
\vspace{-10pt}    
\section{Introduction}

4D Human-Object Interaction (HOI) data are crucial for understanding human behaviors in our three-dimensional world and for numerous applications in computer vision~\cite{Park_2023_CVPR,Yuan_2023_ICCV,zhang2020phosa,Zhang_2023_ICCV,xie2022chore,li2024humanobject,Nam_2024_CVPR}, robotics~\cite{mascaro2023hoiabot,chi2023diffusionpolicy,fishman2022mpinets,lu2024ugg,Wan_2023_unidexgrasppp,2018-TOG-deepMimic}, computer graphics~\cite{starke2019nsm,jiang2024autocharacterscene,hassan2023synthphyint}, and generative AI~\cite{li2023controllablehoisynth,Hu_2024_CVPR_animateanyone,xu2023interdiff,cao2025avatargo}. These applications range from HOI detection and reconstruction to motion generation, robotic learning through demonstration, and even image/video generation. All of these fields rely on 4D HOI data to capture human and object poses, ground-truth geometries, dynamics, forces, and multi-view observations~\cite{xu2024interdreamer,xie2024rhobinchallengereconstructionhuman}. However, the lack of realistic 4D data hampers progress, particularly in scenarios involving multiple objects and detailed manipulations~\cite{gao2024coohoi,Kim2024ParaHomePE}. As both generative and discriminative models advance~\cite{Nam_2024_CVPR, li2023controllablehoisynth, diller2023cghoi, peng2023hoidiff, wu2024humanobjectinteractionhumanlevelinstructions}, the need for high-quality HOI data has become increasingly critical.


Acquiring high-quality 4D HOI data is expensive due to the need for sophisticated motion capture setups and extensive manual data cleaning. Although recent efforts have provided various 4D human-object motion datasets~\cite{bhatnagar22behave,GRAB:2020,wan2022learn,Kim2024ParaHomePE,zhang2022couch,Tian2022SAMPAM,Zhang2024HOIM3CM,zhao2024imhoi,Li2023omomo}, most focus on single-object interactions or lack detailed hand movements. Comprehensive datasets that capture interactions with multiple objects, with full-body and hand motion, remain a gap in the field.


To address this, we introduce \namelongunderlined (\name), a new 4D animation dataset captured from real performance. \name includes 735 curated sequences totaling 7,875 seconds of motion (captured at 30 fps), featuring diverse daily activities and interactions with 63 objects comprising 72 distinct parts. Many scenes involve interactions with multiple objects, such as meal preparation with various utensils, storage organization, and room arrangement. The objects span a wide range of sizes, from small household items like utensils and tools to larger furniture pieces, all modeled based on real-world measurements. All human motions are captured with detailed body and hand movements, accompanied by text annotations.

The acquisition of \name is particularly challenging due to the complexity of recording fine-grained, multi-object interactions. It requires precise calibration, specialized equipment, and extensive post-processing to produce clean, high-quality sequences. By leveraging state-of-the-art techniques, including Large Language Model (LLM)-generated scripts and multi-sensor tracking, we create a dataset with unprecedented detail and fidelity.

Our dataset's distinctive quality stems from our complementary capture approach. To generate diverse motion scripts covering varied daily activities, we use a directorial mindset to design stories and actions, and we introduce a \textit{Scene-Driven LLM Scripting} method to hierarchically generate these scripts. To capture human motion in the presence of frequent object occlusions, we utilize motion capture suits and gloves with electromagnetic field (EMF) technology to track performers, while dual-Kinect RGB-D sensors record object poses. This multi-modal system ensures fidelity in both large-scale movements and fine manipulations, even in occlusion-heavy scenarios.

All sequences undergo rigorous cleaning and independent verification by professional artists, with particular attention to common issues such as foot sliding and object penetration, ensuring clean yet natural movement nuances preserved data for machine learning context. An independent group of artists were also invited to assess the complete dataset's quality from a professional perspective. Moreover, we introduce a set of metrics to evaluate our and other HOI datasets, providing a comprehensive benchmark for human motion, object motion, and interaction quality. 

\name provides a valuable resource for training models in motion generation, robotics, computer vision, and 2D generation. These sequences capture not only physical dynamics but also the logical progression of tasks, making them useful for learning natural action sequences and task planning~\cite{Li2023omomo,mascaro2023hoiabot}. The comprehensiveness of the data set extends its utility in multiple domains: motion generation models can learn natural interaction patterns~\cite{diller2023cghoi,peng2023hoidiff}, robotics researchers can study human manipulation strategies~\cite{mascaro2023hoiabot,2018-TOG-deepMimic,Fu2024MobileAloha}, and computer vision systems can train on accurate 3D ground truth for detection, tracking, and reconstruction~\cite{li2024humanobject,Park_2023_CVPR,zhang2020phosa,xie2022chore}. Image or video generation systems can also use verified motion sequences for content creation and authorization~\cite{hassan2023synthphyint,xu2023interdiff, Kim2016RetargetingHoi}.


The contributions of this work include the following.
\begin{itemize}
    \item A high-fidelity HOI dataset featuring complex, meaningful, and diverse daily interactions with multiple objects at various scales.
    \item A multi-modal capture methodology with Scene-Driven LLM Scripting and multi-sensors setup, preserving subtle interactions even in challenging occlusion scenarios.
    \item A set of quality metrics and benchmarks to evaluate HOI datasets to establish quantitative standards for human motion, object motion, and interaction quality.
\end{itemize}

\section{Related Work}

\noindent\textbf{Human Motion Capturing Technologies.}
Recent advances in \textbf{human pose estimation from cameras}, including monocular RGB and RGB-D setups, have significantly broadened the scope of human motion capture. Early research explored markerless systems~\cite{15Corazza2010MarkerlessMC, 18performance, 24Furukawa2008Dense3M, 25Gall2009MotionCU, 56Kehl2006MarkerlessTO, 90Stoll2011FastAM}, while more recent frameworks such as OpenPose~\cite{Cao2018OpenPoseRM} and DensePose~\cite{Gler2018DensePoseDH} provide robust 2D and 3D joint detection. These camera-based systems are frequently improved using optimization techniques~\cite{Hampali2019HOnnotateAM} or pre-trained models~\cite{zhao2024imhoi, Sener2022Assembly101AL}, which substantially improve tracking accuracy. In parallel, marker-based pose estimation methods have been successfully applied to human-object interaction scenarios~\cite{liu2024taco, AMASS:2019, Loper:SIGASIA:2014}, delivering superior precision in specific contexts. Although these techniques are effective in unconstrained environments, they often encounter limitations when dealing with complex poses or occlusions.

Motion capture suits (mocap) have emerged as a widely adopted tool for capturing high-fidelity human motion across both research and industry applications. Both optical mocap systems and electromagnetic field suits have been employed in dataset collection~\cite{kobayashi2023motion, 38Han2018OnlineOM, 26GarciaHernando2017FirstPersonHA, 19DelPreto2022ActionSenseAM}, offering extensive coverage for more challenging scenarios.

For \textbf{object pose estimation}, RGB-D cameras have become increasingly prevalent in HOI scenarios. Advanced techniques~\cite{Xiang2017PoseCNNAC, Tremblay2018DeepOP, Wen_2024_CVPR_foundationpose, lin2023sam6d} have demonstrated remarkable performance in object detection and localization. In the domain of neural systems, inertial measurement units (IMUs) have been attached to objects to track specific parameters~\cite{zhao2024imhoi, Guzov2022InteractionRT, Zhang2022COUCHTC}.

\noindent\textbf{Human-Object Motion Datasets.}
The field of human-object interaction has witnessed the development of several significant datasets, each addressing different aspects of HOI capture. GRAB~\cite{GRAB:2020} represents one of the first data sets that addresses the full-body human-object interaction; however, it focuses primarily on upper-body interactions and is therefore omitted from our comparison. BEHAVE~\cite{bhatnagar22behave} and OMOMO~\cite{Li2023omomo} present more complex scenarios but lack detailed hand pose information. IMHD~\cite{Zhang2024HOIM3CM} specifically targets highly dynamic human-object interactions such as sports activities. Home~\cite{Kim2024ParaHomePE} and TRUMANs~\cite{jiang2024TRUMANS} investigate human-object interactions within domestic environments, though these scenes tend to be more stationary with limited variance. TACO~\cite{liu2024taco} focused more on capturing ego-centric interactions.
Beyond dedicated data sets on human-object interaction, MIXAMO~\cite{mixamo} provides a comprehensive repository of motion capture data used primarily in character animation and game development. HUMAN3.6M~\cite{human36m} constitutes a large-scale dataset designed for human motion capture, focusing on natural daily activities rather than human-object interactions.

While each of these datasets has significantly advanced the field, all exhibit limitations in capturing the complexity of real-world multi-object interactions. A critical shortcoming is the frequent inaccuracy of hand-object interactions, where hands either appear completely detached from objects or penetrate surfaces by significant margins. Additionally, many existing datasets consist of isolated, purposeless movements that, even with textual annotations, make it difficult to extract meaningful representations of continuous human daily activities. These limitations impede the development of models capable of understanding natural human-object interactions, particularly when involving multiple objects or requiring fine-grained manipulations.

\begin{figure}[tb]
    \centering
    \includegraphics[width=\linewidth]{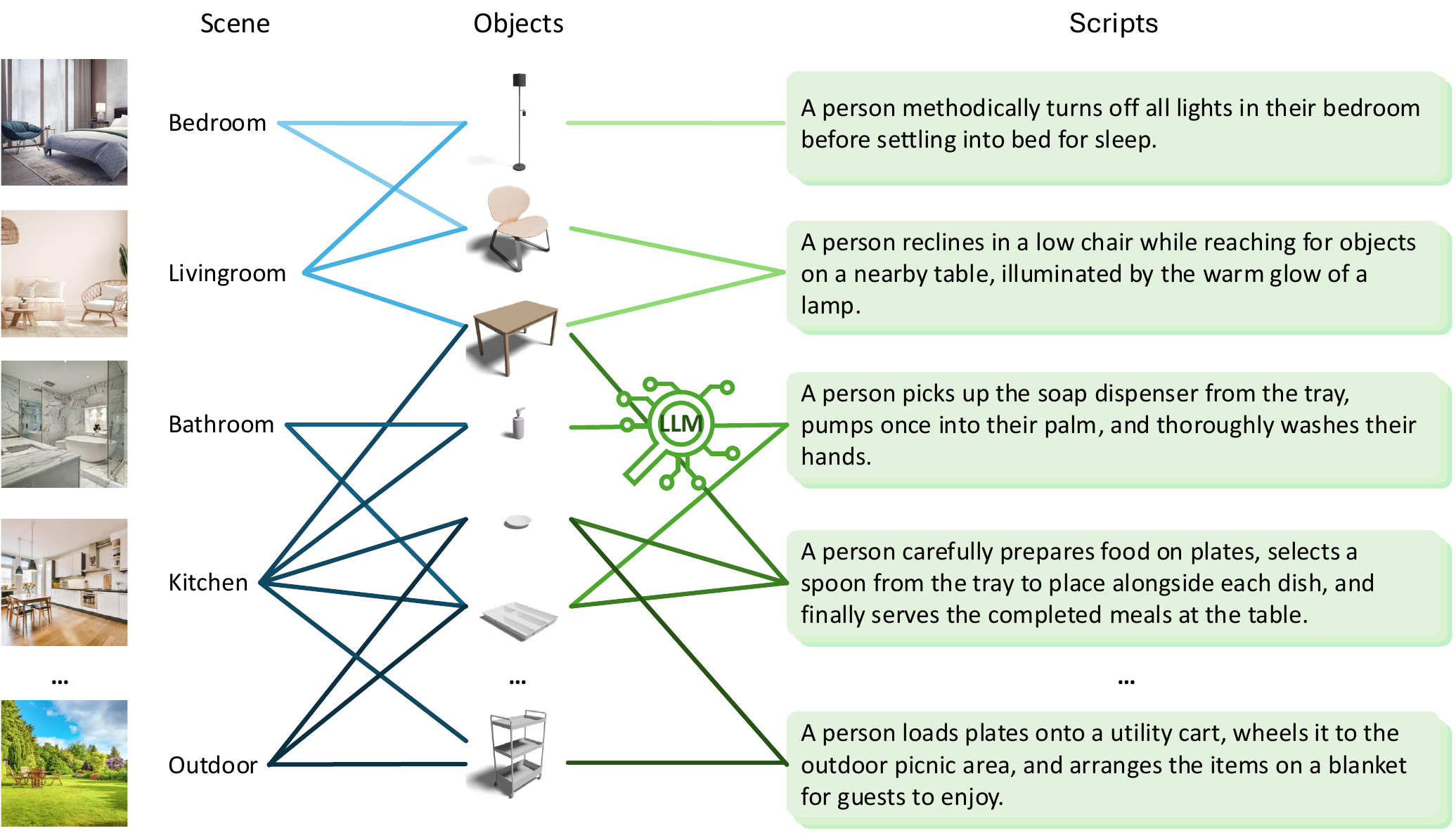}
    \vspace{-10pt}
    \caption{\textbf{Scene-Driven LLM Scripting.} We established target scenes, prepared relevant interaction objects, and then leveraged LLMs to generate detailed action scripts.}
    \label{fig:director_llm}
    \vspace{-15pt}
\end{figure}

\section{Data Collection}
\label{sec:data_collection}
The \name dataset advances human-object interaction research through a comprehensive collection methodology that mirrors cinematic production. Beginning with LLM-generated scripts describing natural daily activities, we carefully selected and modeled common household objects before capturing interactions on a custom motion capture stage equipped with dual Kinect cameras. 

\subsection{Script Development}
\label{sec:script_development}

\begin{figure}
    \centering
    \includegraphics[width=\linewidth]{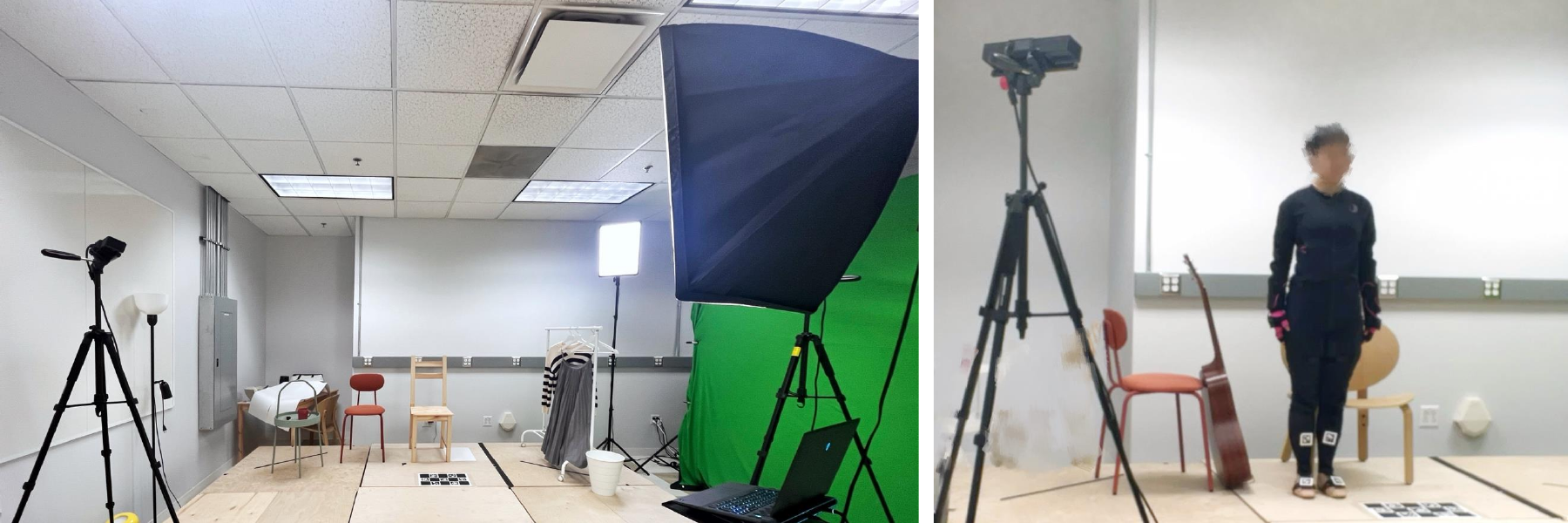}
    \vspace{-20pt}
    \caption{\textbf{Capture environment.} \textit{Left:} Overview of our capturing environment showing two Kinect cameras, stage, lighting, calibration board, and interaction objects. \textit{Right:} Calibration procedure with the performer in a standardized position, enabling precise alignment between mocap suit data and camera coordinates.}
    \label{fig:capture_environment}
    \vspace{-5pt}
\end{figure}

To address the limitations of existing datasets, where interactions often appear arbitrary or disconnected, we develop a systematic approach to create action scripts before capturing a large volume of motion data. Inspired by movie production workflows of grouping actions into scenes, we established a \textit{Scene-Driven LLM Scripting} framework to automate script generation. First, we created conceptual ``rooms'' by logically grouping related objects from our collection. We then provided these groupings to LLMs to generate cohesive interaction sequences within contextual spaces, as illustrated in \cref{fig:director_llm}. This resulted in rich narratives where performers executed purposeful tasks, such as opening a drawer to retrieve an item, arranging objects on a desk, or preparing a meal, thereby ensuring that each motion served a clear function. Further details of the \textit{Scene-Driven LLM Scripting} process are provided in the supplemental materials (Fig. \textcolor{iccvblue}{14}).


\begin{figure}[t]
    \centering
    \includegraphics[width=\linewidth]{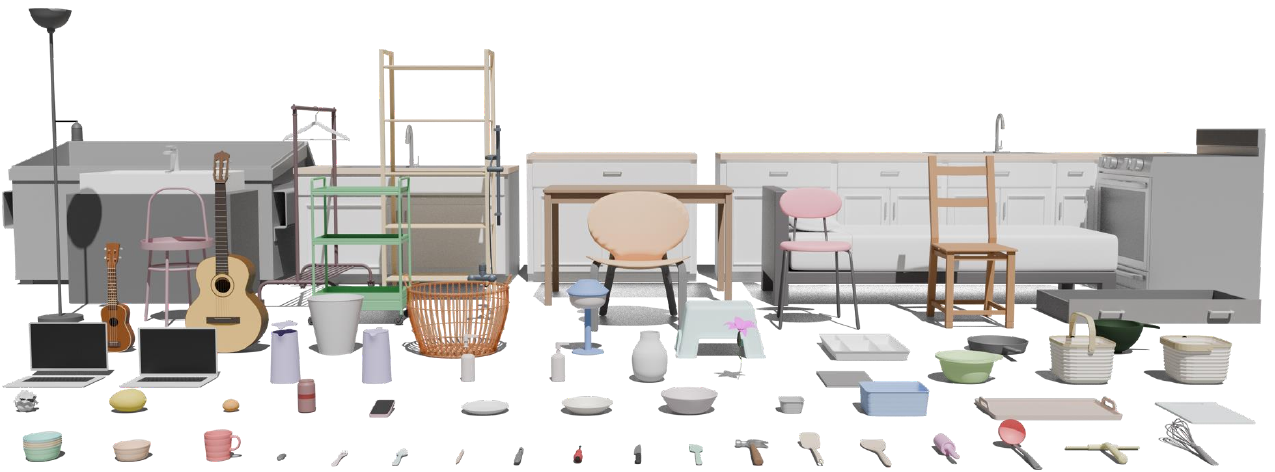}
    \vspace{-20pt}
    \caption{\textbf{3D Meshes.} Artist-modeled objects used in \name.}
    \label{fig:objects}
    \vspace{-10pt}
\end{figure}

\begin{figure*}[t]
    \centering
    \includegraphics[width=\linewidth]{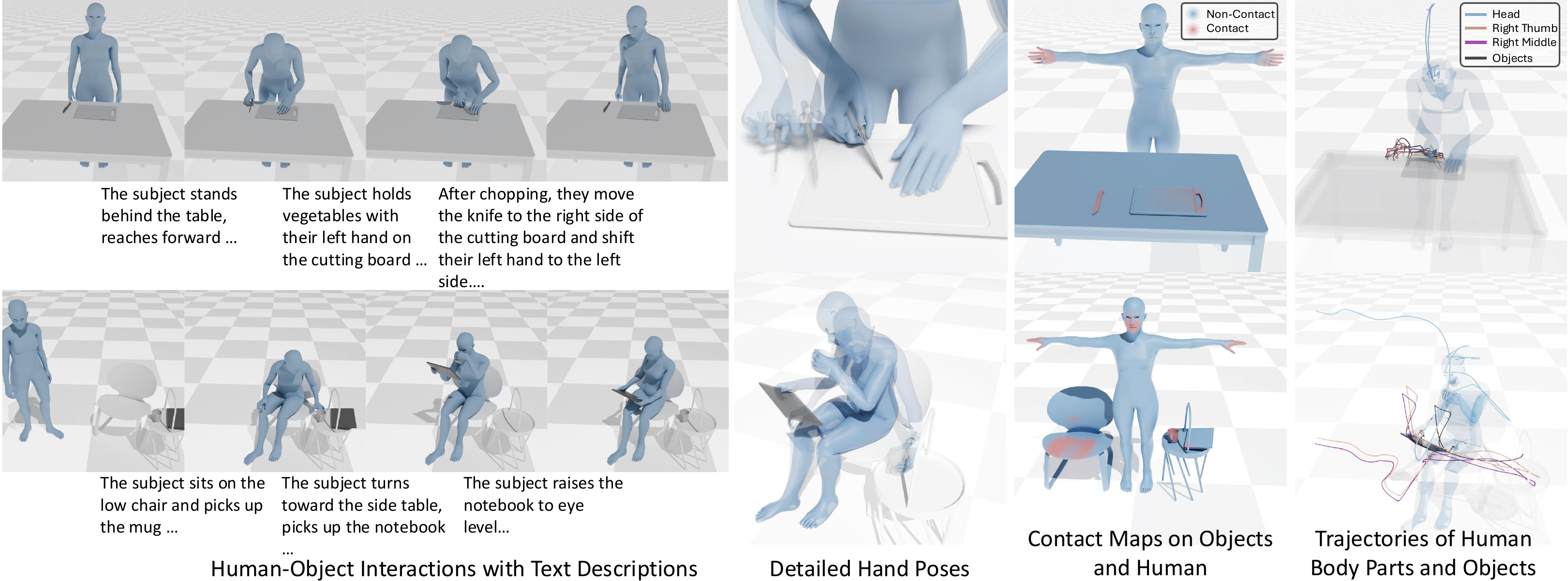}
    \vspace{-15pt}
    \caption{\textbf{\name dataset visualization.} We depict human-object interactions with text descriptions \textit{(left)}, detailed hand poses, and contact maps highlighting interaction areas \textit{(middle)}, and trajectories of human body parts and objects during activities \textit{(right)}. These complementary representations provide comprehensive data for various applications.}
    \label{fig:generation_assets}
    \vspace{-10pt}
\end{figure*}

\subsection{Environment and Capturing}
\label{sec:hardware_setup_and_capturing}

\noindent\textbf{Objects and Humans.}
\name is built on a carefully curated collection of 63 standard household objects, encompassing 72 distinct functional parts (\cref{fig:objects}). Unlike previous datasets relying on 3D scanning, we recruited professional artists to create precise digital models capturing crucial details, including articulated components and graspable surfaces. This ensures geometric accuracy while preserving part-level information essential for realistic interaction. 

Our performer, outfitted with Rokoko smart-suits~\cite{rokoko_smartsuit_pro_ii} and paired gloves, enabled high-fidelity tracking of full-body movements and finger articulations at 30 frames per second. The inertial sensor network provided reliable skeletal tracking, while the specialized gloves captured the fine-grained hand movements essential for natural object manipulation. The skeletal motions are transferred to a neutral human model with the standard Mixamo skeleton rigging~\cite{mixamo}.

\noindent\textbf{Environment Setup.}
To minimize magnetic interference between the Rokoko suit's inertial sensors and metallic structures in the vicinity, we built a customized wooden stage to elevate performers from the floor by 12 inches. Two Kinect cameras were positioned at two corners, maximizing capture volume while minimizing occlusions during complex interactions. Spatial alignment between camera and motion capture systems was achieved using a calibration board to establish a common coordinate system. The dual-computer setup, one managing Kinect camera feeds and object tracking and the other handling Rokoko motion capture data, maintained precise temporal synchronization through UDP commands routed over a dedicated network.


\noindent\textbf{Capturing process.}
We instructed performers to execute the scripted interactions with purpose rather than mechanical precision, maintaining the fidelity required for data analysis while preserving the characteristic fluidity of human motion. This approach was particularly important in capturing complex sequences with multiple objects, where performers might simultaneously engage with several items, \textit{e.g.}, opening a drawer and reaching for an object, or repositioning multiple items on a surface. We captured these nuanced and complex multi-object interactions, including unconscious behaviors like fidgeting hands that characterize authentic human-environment engagement.

\noindent\textbf{Processing the Raw Data.}
The technical processing pipeline addressed two primary challenges: temporal synchronization and spatial alignment between the human motion capture and object tracking data streams. At the beginning of each capture sequence, performers adopted a calibration stance at a predetermined position where the Rokoko system exhibited optimal tracking performance. This position, mapped to the camera coordinate system using our calibration board reference, established a transformation matrix that aligned both coordinate frames.

Object tracking leveraged the dual Kinect camera setup to minimize occlusions. The FoundationPose~\cite{Wen_2024_CVPR_foundationpose} algorithm analyzed the visual data to determine 6DoF poses for each object. To address the limitations of frame-to-frame consistency assumptions during rapid movements, we implemented a dynamic reset mechanism based on mask pixel differences, reinitializing tracking when substantial movement was detected. To further improve the tracking result, we provide object masks by employing SAM2 with strategic human annotations, ensuring tracking consistency across frames where objects might be temporarily occluded.

\subsection{Data Cleaning and Annotation}
\label{sec:data_cleaning_and_annotation}

\noindent\textbf{Multi-stage Quality Assurance.}
Our quality assurance protocol is a two-stage approach combining technical refinement and independent verification. During technical refinement, professional artists refined capture artifacts like drift and tracking errors, ensuring logical consistency in the interactions. During the subsequent verification, an independent team verified the sequences for natural and plausible human-object interactions, addressing issues such as joint jitter and foot sliding. We iterated these two stages till all quality standards were met, ensuring fidelity to natural movements and interactions.

\begin{table*}[tb]
    \centering
    \resizebox{\linewidth}{!}{
    \begin{tabular}{lC{0.2cm}cccccC{0.2cm}cC{0.2cm}ccc}
    \toprule
        Dataset && \multicolumn{5}{c}{Human Motion} && Object Motion && \multicolumn{3}{c}{Contact} \\
        \cmidrule{3-7}\cmidrule{9-9}\cmidrule{11-13}
        && Foot Sliding & Jerk & MSNR & Coherence & Diversity && Jerk && Penetration & Contact & State \\
        && ($cm$) $\downarrow$ & ($m/s^3$) $\downarrow$ & (dB) $\rightarrow$ & $\uparrow$ & $\uparrow$ && ($m/s^3$) $\downarrow$ && ($cm$) $\downarrow$ & Entropy $\uparrow$ & Consistency $\uparrow$ \\ \midrule
        BEHAVE~\cite{bhatnagar22behave}  && 4.556 & 4.08 & 5.51 & 0.533 & 0.966 && 10.40 && 0.0606 & \underline{2.2915} & 0.0667 \\
        OMOMO~\cite{Li2023omomo} && 2.130 & 15.10 & \underline{12.37} & \underline{0.619} & \underline{0.978} && 27.40 && \underline{0.0602} & 1.9468 & 0.4837 \\
        IMHD~\cite{zhao2024imhoi} && \underline{1.474} & \textbf{1.14} & 14.20 & 0.554 & 0.951 && 24.06 && 0.1172 & \textbf{2.4265} & 0.2411 \\
        ParaHome~\cite{Kim2024ParaHomePE} && 3.008 & 9.19 & 1.82 & 0.592 & \textbf{0.980} && \textbf{0.08} && 0.2167 & 1.0254 & \textbf{0.6815} \\
        \midrule
        \name && \textbf{0.958} & \underline{1.87} & \textbf{9.42} & \textbf{0.653} & 0.956 && \underline{1.13} && \textbf{0.0068} & 1.4587 & \underline{0.5061} \\ \midrule
        Mixamo && 3.184 & 8.14 & 10.88 & 0.616 & 0.958 && - && - & - & - \\
        \bottomrule
    \end{tabular}
    }
    \vspace{-5pt}
    \caption{\textbf{Quantitative evaluation across human-object interaction datasets.} Metrics defined in Appendix \textcolor{iccvblue}{A.2.1} should be interpreted holistically, as optimal values depend on specific applications. The table includes two additional statistical indicators that provide context for understanding dataset characteristics. Bold indicates \textbf{best}, underline indicates \underline{second-best}. $\uparrow$: higher values are better, $\downarrow$: lower values are better, and $\rightarrow$: values closer to Mixamo are better.}
    \label{tab:dataset_quantitative}
    \vspace{-15pt}
\end{table*}

\noindent\textbf{Textual Annotation.}
We invited an independent group to provide textual descriptions for each sequence based on the actual performance. These annotations included three elements: (1) a concise title highlighting the sequence's main goal with details on subtle differences, (2) a short script providing a complete yet brief description of the motion and interaction in the scene, and (3) a detailed long script elaborating on specific motions and interactions throughout the sequence. These multi-level textual annotations enhance the dataset's utility for applications requiring both visual and semantic understanding of human-object interactions.

This comprehensive approach with script generation, capture, processing, quality control, and annotation, resulted in a dataset that captures both the mechanics of object manipulations and the purposeful sequences in which these manipulations naturally occur. \name provides researchers with data reflecting how humans chain multiple actions together to achieve higher-level goals, enabling advances in human behavior prediction~\cite{hu24hoimotion}, robotic learning from demonstration~\cite{DelPreto2020HelpingRobotsLearn,Fu2024MobileAloha}, virtual character animation~\cite{Kovar2002MotionGraph,jiang2024motiongpt,2018-TOG-deepMimic}, and augmented reality applications~\cite{Kim2016RetargetingHoi}.

\section{Dataset Analysis}


This section elaborates on the quantitative and qualitative evaluations of the motion quality and compares \name against existing datasets: BEHAVE~\cite{bhatnagar22behave}, OMOMO~\cite{Li2023omomo}, IMHD~\cite{zhao2024imhoi}, ParaHome~\cite{Kim2024ParaHomePE} and GRAB~\cite{GRAB:2020}.

\subsection{Quantitative Evaluation}

We evaluate human motion, object motion, and human-object interaction using metrics such as foot sliding, jerk, penetration, contact entropy, and state consistency. These metrics offer insights into motion quality, interaction realism, and the diversity of interaction states.
Additionally, we introduce a new metric, \textbf{Motion Signal-to-Noise Ratio (MSNR)}, to assess the quality of motion relative to noise in the dataset. MSNR evaluates motion quality using the signal-to-noise ratio (SNR)~\cite{Shannon1949CommunicationIT} of joint kinematics. Higher SNR values indicate smoother motion, though excessive smoothing may result in loss of important details. We use Mixamo, an industry-standard motion capture dataset cleaned by artists, as the baseline for human motion quality. Datasets with MSNR values closer to Mixamo's indicate comparable motion quality.
Further details and metric formulations are provided in Appendix \textcolor{iccvblue}{A.2.1}. 


\noindent\textbf{Comparison on Human and Object Motion.}
\name demonstrates superior performance in several key motion quality metrics. The data set exhibits the lowest foot sliding among all datasets compared, significantly outperforming established datasets like BEHAVE~\cite{bhatnagar22behave} and ParaHome~\cite{Kim2024ParaHomePE}. This improvement can be attributed to our meticulous motion capture process and rigorous artist-led quality control. The low jerk values for human motion indicate smooth and natural movements, second only to IMHD \cite{zhao2024imhoi}, whose fast movements are more likely to have similar acceleration in a sequence. While the Mixamo dataset shows higher foot sliding and jerk, it is important to note that Mixamo contains specialized movements like street dancing, which inherently involves more dynamic motions that increase these metrics compared to typical HOI scenarios.

\name achieves 9.42 dB in Motion SNR, approaching Mixamo's reference value. This slightly lower SNR compared to IMHD and OMOMO stems from \name's complex interactions with detailed hand poses, which introduce higher frequency components often interpreted as ``noise''. Notably, OMOMO's combination of high SNR with high jerk values suggests clean signals that still contain abrupt motion changes, a phenomenon meriting future investigation. \name's high coherence demonstrates consistent, targeted motions while maintaining competitive diversity, especially compared to Mixamo. The unusually high diversity scores of other datasets may indicate excessive noise rather than true motion variety, artificially inflating their entropy measurements.

In object motion, \name demonstrates a notably low jerk, indicating realistic object manipulation, unlike the high values in OMOMO and IMHD. ParaHome's extremely low object jerk reflects that the objects in their long sequences are mostly static and barely interact with humans.

\noindent\textbf{Comparison on Contact Quality.}
\name excels in contact quality metrics, achieving the lowest penetration among all datasets despite including detailed hand poses. This order of magnitude improvement over BEHAVE and OMOMO demonstrates our exceptional precision in capturing human-object spatial relationships, which is crucial for physically plausible interaction models.
The contact entropy for \name shows a balanced distribution between contact states, more diverse than ParaHome but more focused than the potentially noisy patterns in IMHD and BEHAVE, suggesting meaningful interactions without excessive fluctuations.
For state consistency, \name strikes a balance between the highly consistent but potentially oversimplified ParaHome and the less consistent BEHAVE, maintaining realistic transitions while avoiding rapid fluctuations that might indicate tracking errors.

Overall, \name combines the detailed hand articulation with superior metrics in foot sliding, smoothness of object motion, and minimal penetration, making it valuable for applications requiring physically accurate human-object interactions with natural motion.




\begin{table}[tb]
\resizebox{\linewidth}{!}{
\begin{tabular}{l<{\hspace{-10pt}}c<{\hspace{-10pt}}c<{\hspace{-10pt}}c<{\hspace{-5pt}}c<{\hspace{-5pt}}c<{\hspace{-10pt}}ccc}
\toprule
    Dataset & \# hours & \# subj. & \# obj.  & hand & body & max. obj. & setup \\
    \midrule
    GRAB~\cite{GRAB:2020} & 3.8 & 10 & 51  & \cmark & \cmark & 1 & standing \\
    BEHAVE~\cite{bhatnagar22behave} & 4.2 & 8 & 20  & \xmark & \cmark & 1 & portable \\
    InterCap~\cite{Huang2022InterCapJM} & 0.6 & 10 & 10  & \cmark & \cmark & 1 & portable \\
    OMOMO~\cite{Li2023omomo} & 10.1 & 17 & 15  & \xmark & \cmark & 1 & portable \\
    FHPA~\cite{GarciaHernando2017FirstPersonHA} & 0.9 & 6 & 26  & \cmark & \xmark & 1 & room \\
    HOI4D~\cite{Liu2022HOI4DA4} & 22.2 & 9 & 800  & \cmark & \xmark & 1 & room \\
    Chairs~\cite{Jiang2022FullBodyAH} & 16.2 & 46 & 70  & \cmark & \cmark & 1 & standing \\
    ARCTIC~\cite{fan2023arctic} & 1.2 & 10 & 11  & \cmark & \cmark & 1 & standing \\
    NeuralDome~\cite{zhang2023neuraldome} & 4.6 & 10 & 23 & \cmark & \cmark & 1 & standing \\
    TRUMANS~\cite{jiang2024TRUMANS} & 15 & 7 & 20 & \cmark & \cmark & - (proxies) & room \\
    ParaHome~\cite{Kim2024ParaHomePE} & 8.1 & 38 & 22  & \cmark & \cmark & 22 & room \\ \midrule
    HUMOTO & 2.2 & 1 & 63 & \cmark & \cmark & 15 & scene \\
    \bottomrule
\end{tabular}
}
\vspace{-5pt}
\caption{\textbf{Dataset statistics.} We provide details on the total durations, number of subjects, objects, presence of hand and body data, maximum objects in scene, and data collection setup styles.}
\vspace{-10pt}
\label{tab:statistics}
\end{table}

\subsection{Qualitative Evaluation}
The quantitative results are influenced by features of the datasets that do not necessarily represent quality issues. Therefore, they should be interpreted holistically rather than in isolation, as their values are influenced by multiple factors, including motion and interaction complexity. Thus, we also provide qualitative evaluations.

\subsubsection{Visual Quality}
\begin{figure}[t]
    \centering
    \includegraphics[width=\linewidth]{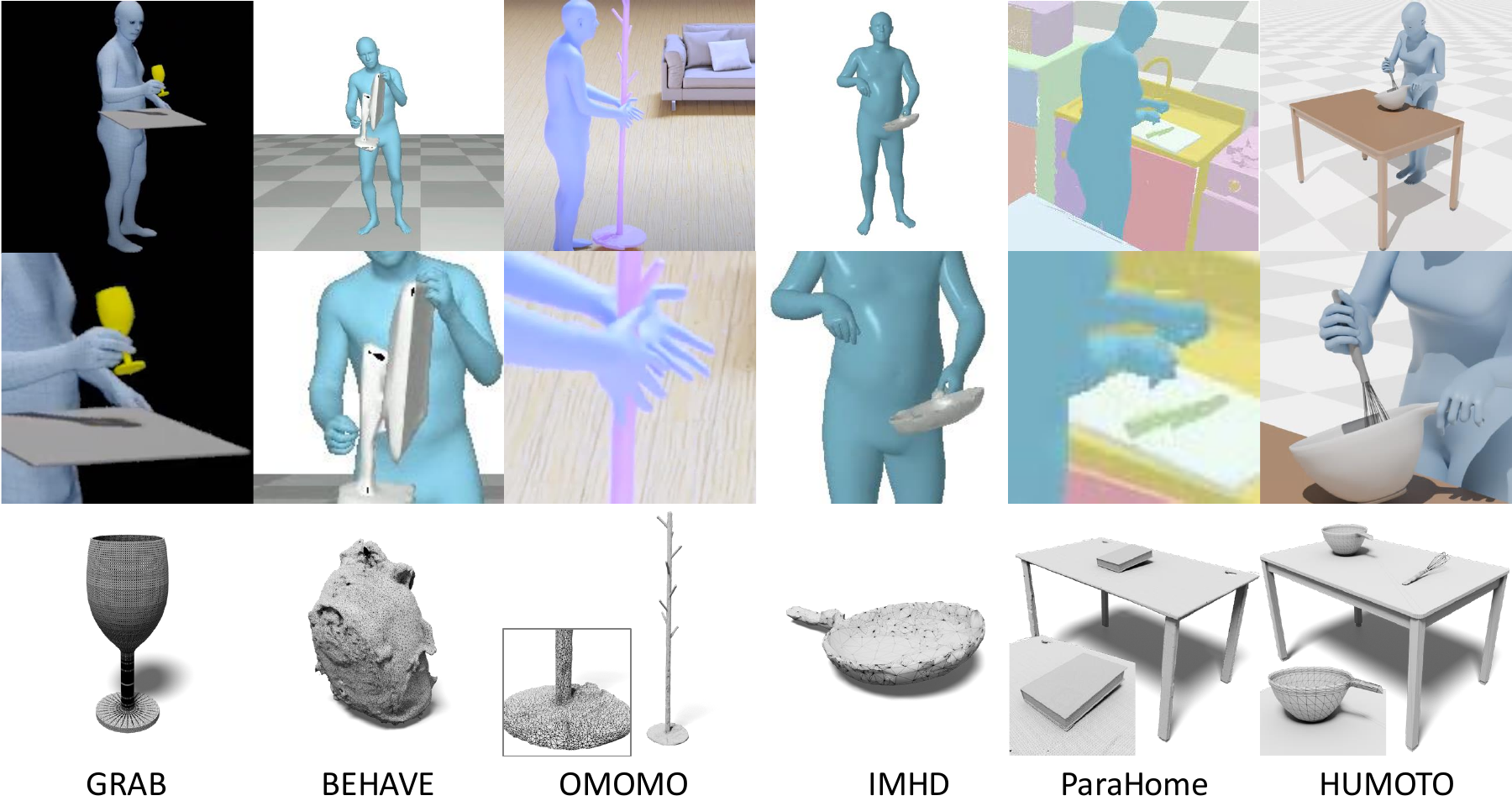}
    \vspace{-15pt}
    \caption{\textbf{Quality comparison.} We compare different datasets on motion dynamics, hand pose accuracy, and object meshes.}
    \label{fig:dataset_comparisons}
    \vspace{-15pt}
\end{figure}

We present a visual quality comparison in \cref{fig:dataset_comparisons}. While BEHAVE and OMOMO use only standard hand templates without detailed finger poses, IMHD offers finer hand modeling but exhibits significant penetration in several scenes. ParaHome provides relatively flexible hand motion, though their capture method (attaching tags on hands) interferes with natural movement, resulting in frequent clenched hand poses throughout the dataset. \name demonstrates superior hand pose quality, particularly during interactions. We also compare object mesh quality across datasets. Objects from prior datasets show noise artifacts due to 3D scanning limitations, while our object modeling pipeline produces clean, accurate representations.


\subsubsection{Perceptual Study}\label{sec:perceptual_study}
To complement our quantitative analysis, we conducted a human perceptual study evaluating \name against existing HOI datasets through absolute quality assessment and direct pairwise comparison. We report the results of an online study taken by 26 participants, comprising students and researchers specializing in computational human motion.

\noindent\paragraph{Absolute Quality Assessment.}
Participants rated randomly selected videos from \name, BEHAVE, IMHD, OMOMO, and ParaHome on a 5-point Likert scale. \name achieved the highest scores in all categories: human motion (4.79$\pm$0.49), with 82\% giving maximum scores), object motion (4.88$\pm$0.36), interaction quality (4.75$\pm$0.57), and overall quality (4.78$\pm$0.43). These scores significantly outperformed all comparison datasets, with the most notable difference in interaction quality, where BEHAVE scored only 2.48$\pm$1.05 and even recent datasets like IMHD (3.94$\pm$1.04) lagged considerably.




\noindent\paragraph{Pairwise Comparison.}
In this study, participants directly compared \name against other datasets showing the same interaction tasks. The results strongly favored \name in all dimensions, with 96\% preferring \name over BEHAVE for overall quality. Even against newer datasets, \name was consistently preferred: 46\% versus IMHD (with 50\% rating both equally good), 65\% versus OMOMO (28\% ties), and 82\% versus ParaHome (15\% ties). For interaction quality specifically, \name outperformed BEHAVE (94\% preference), OMOMO (65\%), and ParaHome (67\%), while against IMHD, \name was preferred by 38\% and rated equally good by 46\%.

These results demonstrate the superior quality of \name in both absolute ratings and direct comparisons, particularly for interaction quality and overall performance. Details are provided in Appendix \textcolor{iccvblue}{A.3}. 

\section{Discussions}
Building upon the novel script generation pipeline, the multi-sensors motion capture system, and the rigorous quality control described in \cref{sec:data_collection}, the \name dataset provides not merely the detailed mechanics of object manipulation but the purposeful sequences in which these manipulations naturally occur. \name offers exceptional value for a wide range of research and applications, of which we highlight some below.





\begin{figure}
    \centering
    \includegraphics[width=\linewidth]{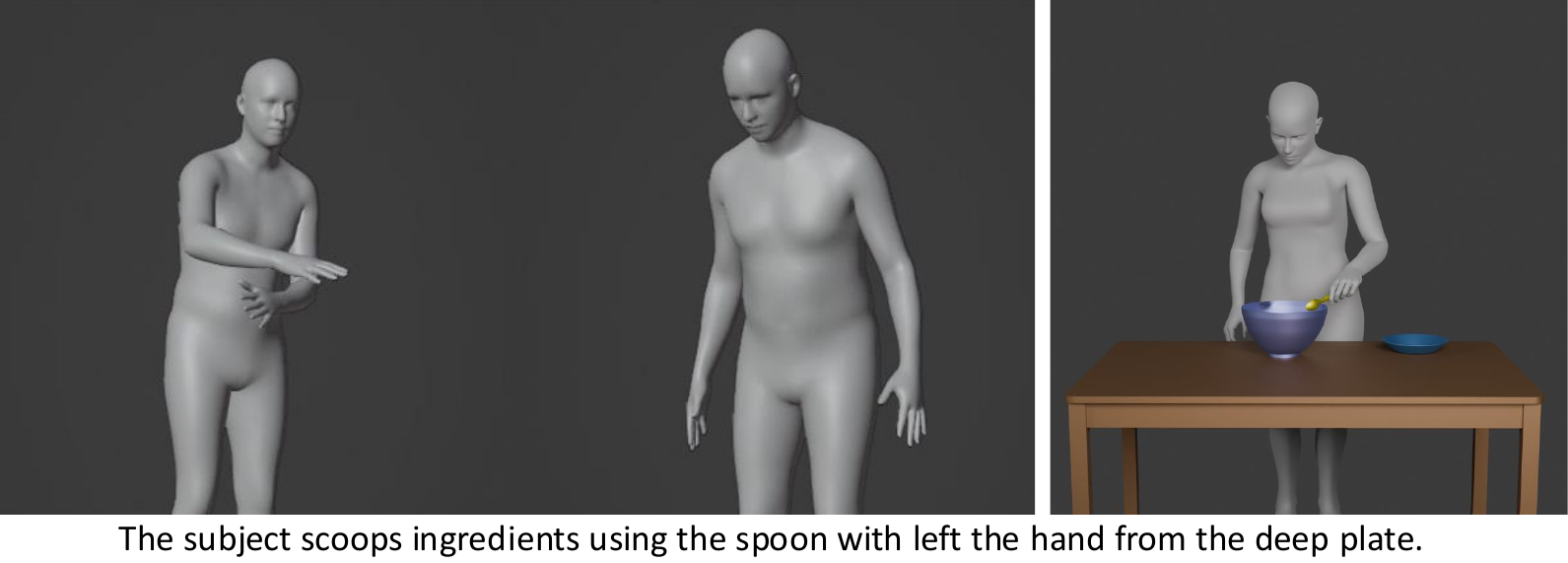}
    \vspace{-15pt}
    \caption{\textbf{Motion Generation by MotionGPT~\cite{jiang2024motiongpt}.} \textit{Left:} Motion generated from the short scrip. \textit{Mid:} Motion generated from the long script. \textit{Right:} Motion with same text annotation from \name dataset.}
    \label{fig:motion_gpt2}
    \vspace{-10pt}
\end{figure}

\noindent\paragraph{Human-Object Interaction and Motion Generation.}
\label{sec:app:interaction_motion_generation}
Our dataset supports the development of generative models that can translate textual descriptions (\textit{e.g.}, ``pick up the coffee mug and drink from it'') into realistic interaction sequences. The diversity of objects and interactions in \name provides rich supervision for text-conditioned motion synthesis. Our dataset is challenging as state-of-the-art human-object interaction models do not have the ability to generate interaction motion on multiple objects. To show this, we test MotionGPT~\cite{jiang2024motiongpt} with \name prompts in \cref{fig:motion_gpt2}. It appears that the model can generate a few reasonable motions based on the more abstract description, but fails to faithfully generate more fine-grained motions compared to the captured ground truth \name motions. This experiment demonstrates that state-of-the-art motion generation methods, despite being trained with large-scale datasets such as AMASS\cite{AMASS:2019} and HumanML3D \cite{Guo_2022_CVPR}, still struggle with generating detailed human-object interaction. \name is designed to fill this gap.

\vspace{-8pt}

\begin{figure}[tb]
    \centering
    \includegraphics[width=\linewidth]{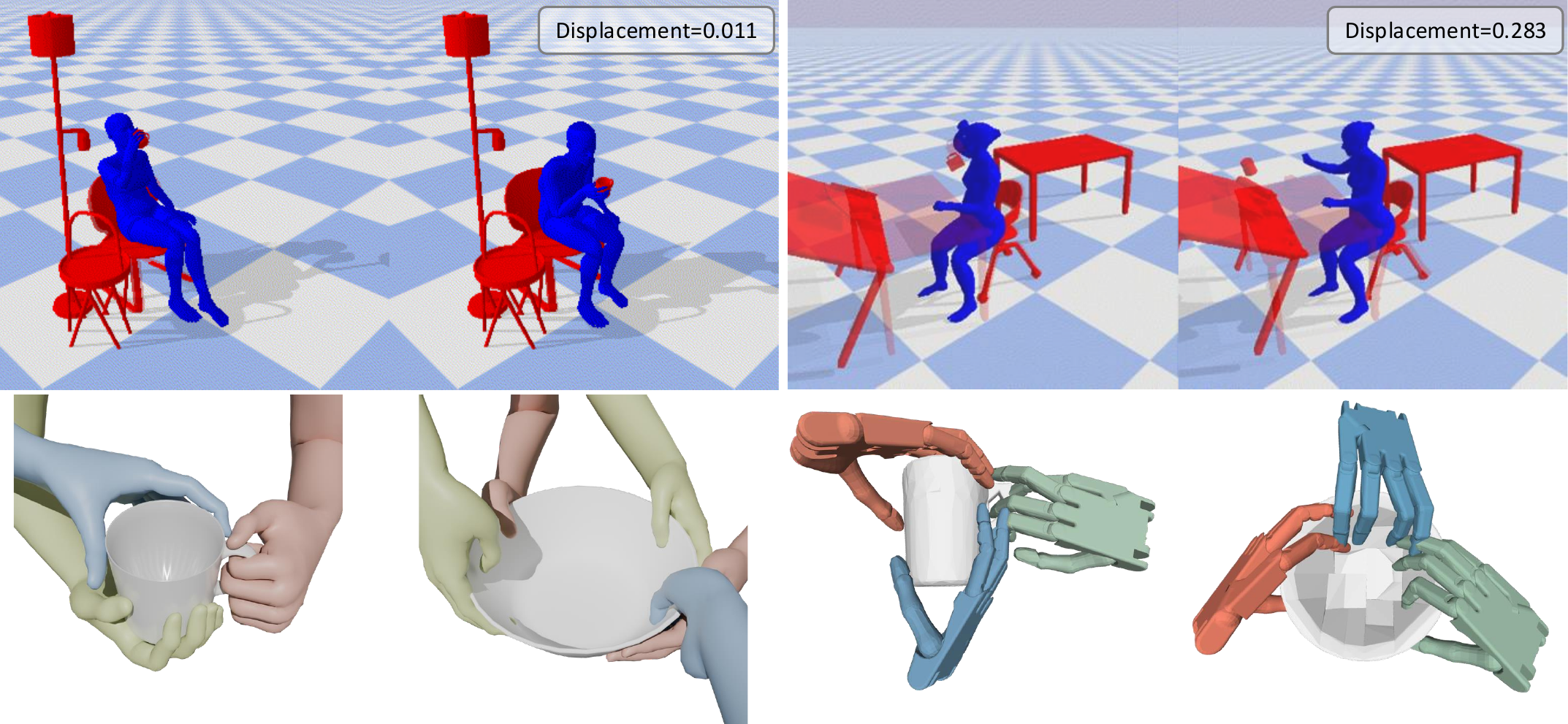}
    \vspace{-18pt}
    \caption{\textbf{Data for Robotics.} \textit{Top:} Two simulator visualizations showing human sitting and holding mug. \name \textit{(left)} displays minimal displacement, while ParaHome \textit{(right)} shows significant object displacement during identical actions. \textit{Bottom:} Hand manipulation comparison between \name \textit{(left)} and simulated robotic grasps from DexGraspNet \textit{(right)}.}
    \vspace{-10pt}
    \label{fig:robot}
\end{figure}

\noindent\paragraph{Robotics and Embodied AI.}
\label{sec:app:robotics}
The precision and diversity of interactions in \name make it particularly valuable for robotics research. To demonstrate the capability of our data, we use PyBullet~\cite{coumans2016pybullet} to compare \name with Parahome~\cite{Kim2024ParaHomePE} in simulation settings. After weighting our objects and assigning similar mass to the ParaHome dataset, we use CoACD~\cite{wei2022coacd} to obtain convex shapes for simulation. Overlaying the final frame on the first (\cref{fig:robot}, Top) reveals significantly smaller object displacement in our dataset compared to Parahome, where interacted objects show substantial movement. Grasp synthesis, a popular robotic research topic~\cite{Wan_2023_unidexgrasppp,Wang2022DexGraspNetAL,lu2024ugg}, usually relies on simulated data that, despite passing simulator validation, often produces unnatural (\textit{e.g.}, blue hand on mug bottom) or functionally unreasonable grasps (\textit{e.g.}, fingers inside bowls). Comparing similar object grasps from \name with those from DexGraspNet~\cite{Wang2022DexGraspNetAL} in \cref{fig:robot} (Bottom) shows that our hand poses are more natural and aligned with daily usage. Additionally, \name's task-oriented motion data can help robot learning systems develop capabilities directly transferable to real-world household assistance scenarios rather than simple interaction primitives.
\vspace{-8pt}

\begin{figure*}[tb]
    \centering
    \includegraphics[width=\linewidth]{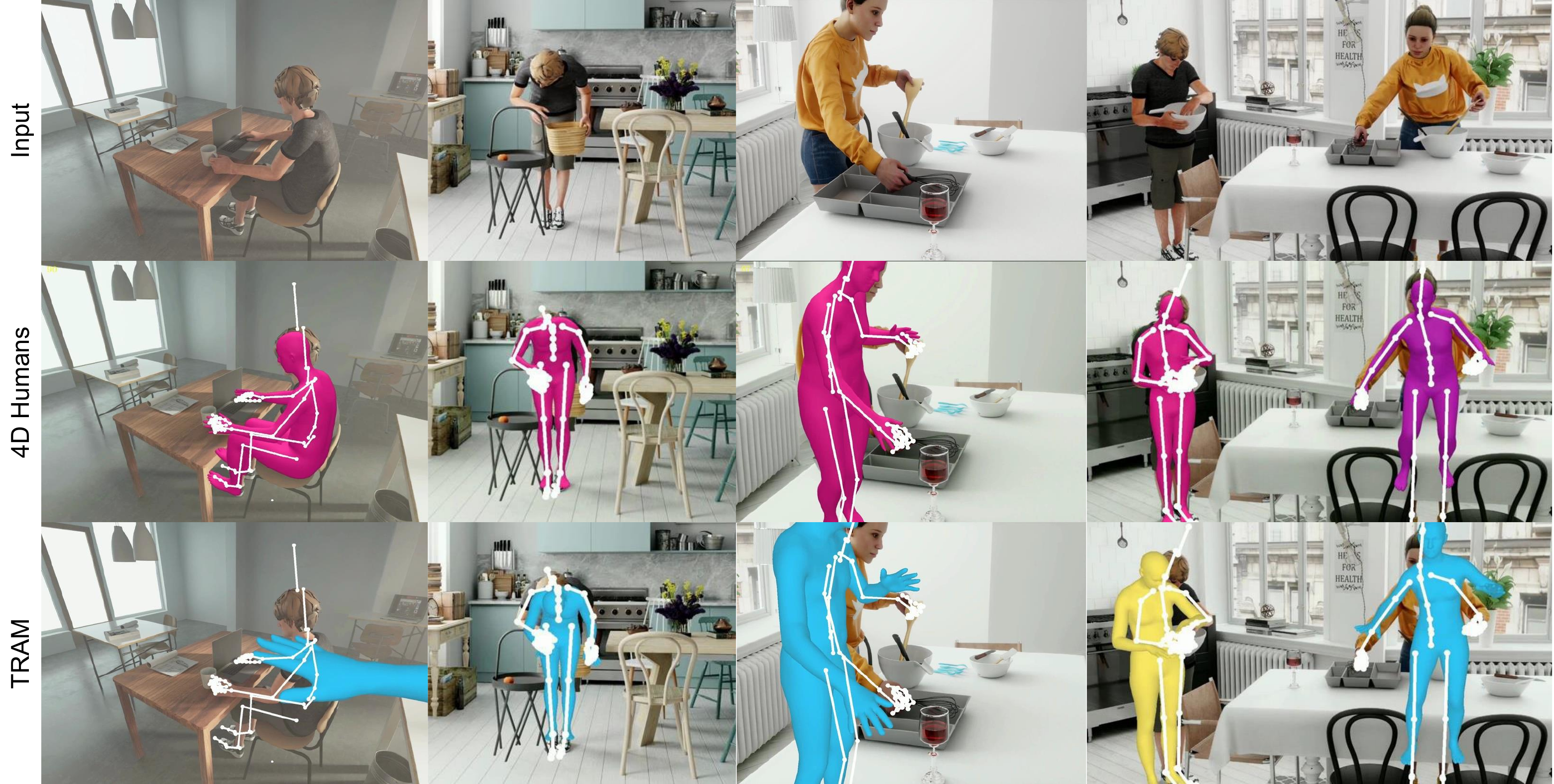}
    \vspace{-15pt}
    \caption{\textbf{Human motion and pose estimation results on \name.} Comparison between 4D Humans~\cite{goel20234dhumans} \textit{(Mid)} and TRAM~\cite{Wang2024TRAMGT} \textit{(Bottom)} on rendered images, showing estimated meshes (colored) against ground truth skeleton (white).}
    \label{fig:pose_estimation}
\end{figure*}

\noindent\paragraph{Pose Estimation in Challenging Scenarios.}
\label{sec:app:pose_estimation}
State-of-the-art human pose estimation methods continue to face challenges in complex interaction scenarios. \name provides precise ground truth for these difficult cases with detailed hand articulation, particularly where objects partially occlude parts of the human body, creating ideal training data for models that must infer joint positions despite visual obstruction. \cref{fig:pose_estimation} demonstrates how even the leading motion and pose estimation models, 4D Humans~\cite{goel20234dhumans} and TRAM~\cite{Wang2024TRAMGT} struggle to predict correct poses from the renderings of our dataset. Additionally, none of these methods incorporates the hand pose estimation capabilities.

\begin{figure*}[tb]
    \centering
    \includegraphics[width=\linewidth]{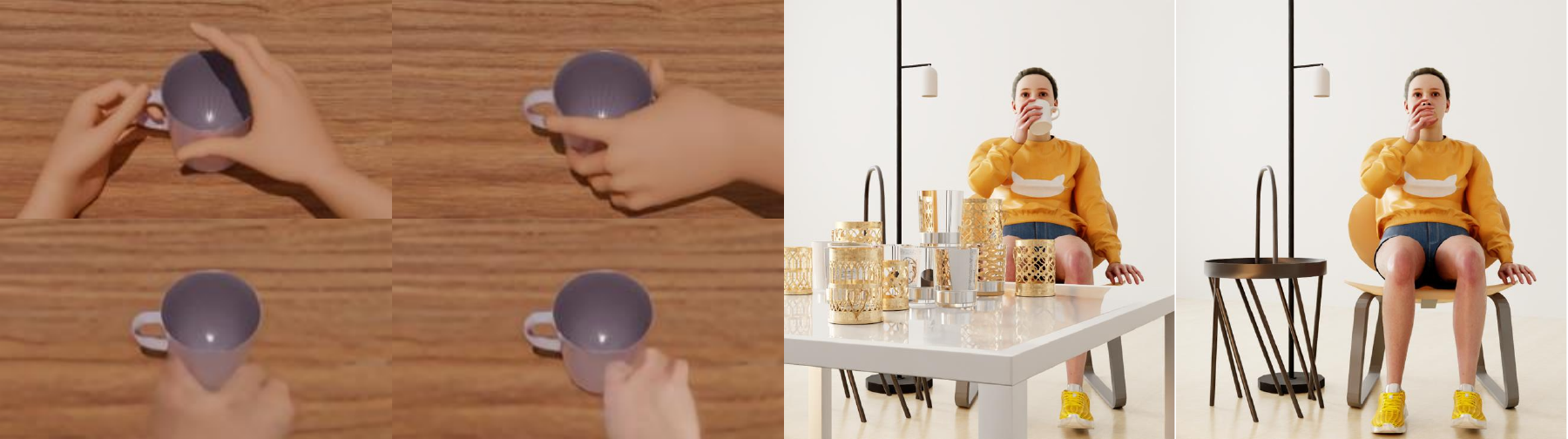}
    \vspace{-15pt}
    \caption{\textbf{Image editing.} \textit{Left:} Hand-object interaction image generation conditioned on a mug. Recent work Affordance Diffusion \textit{(bottom row)} produces physically implausible interactions with imprecise hand positioning, while \name can provide renderings \textit{(top row)} of realistic hand placements at various positions.  \textit{Right:} Our dataset can also be used to provide renderings of object addition and removal, capturing differences in shadows and reflections, and facilitating authorized human-in-scene generative model training. }
    \label{fig:image_editing}
    \vspace{-5pt}
\end{figure*}

\vspace{-8pt}

\noindent\paragraph{Authorized 2D Generation.} \label{sec:app:image_editing} Generating realistic images and videos often requires data that are difficult to capture, such as different viewpoints, object manipulation, or lighting changes. HUMOTO provides rich, human-involved scene data to simulate object addition/removal, reveal occluded areas, and capture lighting and shadow effects (\cref{fig:image_editing}, right). Existing 2D models, like Affordance Diffusion~\cite{ye2023affordance}, often produce artifacts such as distorted hands and blurry poses (\cref{fig:image_editing}, left bottom). HUMOTO offers high-quality, realistic renderings of complex human-object interactions, enabling more accurate training for human-object interaction models~\cite{xue2024hoiswap,ye2023affordance}.

\section{Conclusion and Limitations}
In this work, we present \name, a comprehensive dataset of human-object interactions with detailed and accurate hand motion, and a dedicated scene-driven LLM scripting method to hierarchically design interaction scripts.

Despite \name's advancements, it has some limitations. 
First, due to motion capture suit size constraints, our dataset includes only a single performer, which may introduce a bias toward a particular human body shape and movement style. 
Second, the dataset preparation process required considerable manual cleaning and refinement of the captured motion data. While such manual intervention ensures high-quality data, it represents a significant resource investment. To mitigate this challenge in future work, more advanced and robust pose estimation methods are needed. We hope that \name can serve as a foundational training set for developing such automated techniques, ultimately reducing the manual effort required for high-fidelity human-object interaction data collection.


\section*{Acknowledgement}
The authors would like to thank Jimei Yang for his initial discussions on script development and Nathan Carr for designing and constructing the wooden stage for motion capture. We also thank Ziwen Chen for testing the MoCap suit, and Yuan Yao, Hanwen Jiang, Sanghyun Son, Shoubin Yu for helping us install the objects. Q. Huang acknowledges the support of NSF-2047677, NSF-2413161, NSF-2504906, NSF-2515626, and GIFTs from Adobe and Google. We would like to thank UT GenAI Center for generous support of GPU hours.

\clearpage
\appendix

\section{Dataset Analysis Details}
\subsection{Statistics}
\begin{figure*}[tp]
    \centering
    \includegraphics[width=0.677\linewidth]{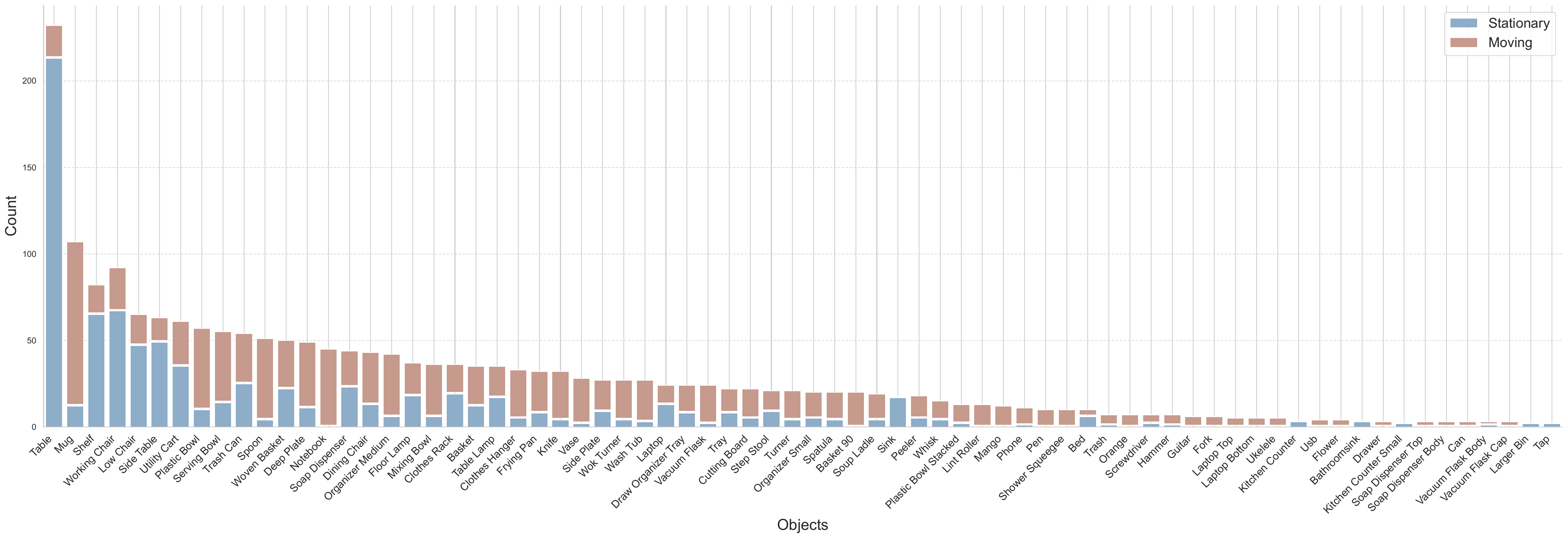}
    \includegraphics[width=0.31\linewidth]{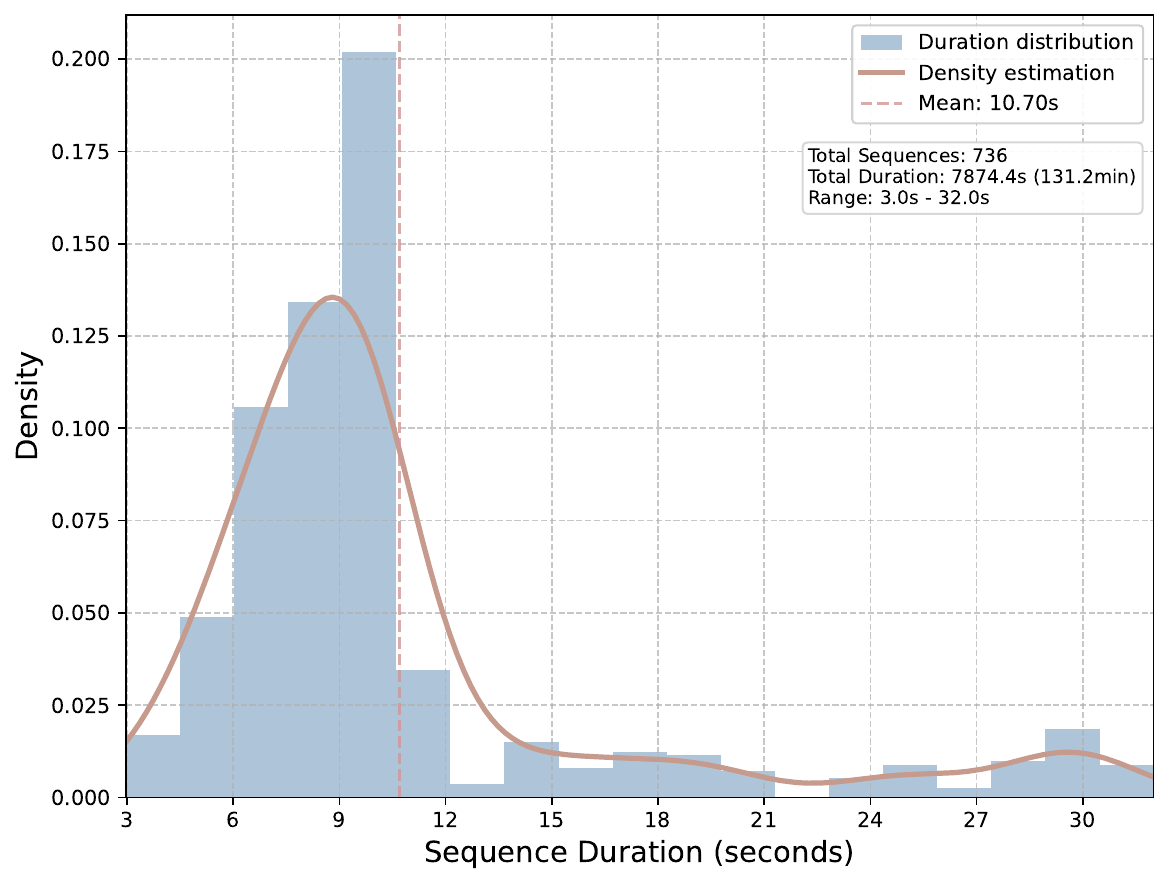}
    \vspace{-8pt}
    \caption{\textbf{Dataset statistics.} \textit{Left:} Object occurrence frequency by motion type (stationary vs. moving). \textit{Right:} Sequence duration distribution across the dataset.}
    \label{fig:statistics}
    \vspace{-10pt}
\end{figure*}
We provide a visualization of statistics of the objects that occur in our dataset and the sequence duration distribution in \cref{fig:statistics}. A detailed comparison of actions, objects and scenes between existing dataset statistics is included in the supp. material.


\subsection{Metrics}
\subsubsection{Metrics Overview}
\label{sec:metrics}
To quantitatively assess the quality of our \name dataset compares to others, we define the following metrics that capture different aspects of motion naturalness and interaction accuracy.

For human and object motion: 
\textbf{Foot sliding} measures unnatural horizontal movement during ground contact. For foot joints below a height threshold, we calculate horizontal displacement with a weighting function that decreases as joints lift from the ground. Lower values typically indicate more natural motion.
\textbf{Jerk} quantifies motion smoothness by measuring the rate of change of acceleration. Lower jerk represents smoother motions. 
\textbf{Motion Signal-to-Noise Ratio (MSNR)} evaluates motion quality through the SNR of joint kinematics. Higher SNR indicates smoother motion, though overly smoothed signals may lose important details. 
\textbf{Coherence} quantifies motion consistency by measuring pose cluster compactness. Values approaching 1 indicate highly consistent movement patterns with minimal deviation. 
\textbf{Diversity} measures variety of motion patterns using normalized Shannon entropy across pose clusters. Higher values indicate a wider range of motion patterns, though this may potentially identify jitter as diversity.

For interaction quality:
\textbf{Penetration} assesses the physical plausibility of human-object interactions by measuring object intrusion into the human mesh. Lower values indicate more physically plausible interactions.
\textbf{Contact entropy} quantifies the diversity of interaction states and transitions. Higher values indicate more diverse and complex interactions with a balanced distribution of contact behaviors.
\textbf{State consistency} measures the temporal stability of interaction states, rewarding smooth contacts while penalizing rapid fluctuations. Higher scores indicate more consistent interaction states with fewer changes.

Jerk is computed for both human and object motion. Foot sliding, MSNR, Coherence, and Diversity apply only to human motion. Penetration, contact entropy, and state consistency evaluate human-object interaction quality. These metrics are influenced by features of the dataset that do not necessarily represent quality issues. Therefore, they should be interpreted holistically rather than in isolation, as their values are influenced by multiple factors including motion and interaction complexity. A complete definition of metrics is provided in \cref{sec:metrics_formulation}.

\subsubsection{Metrics Formulation}
\label{sec:metrics_formulation}

\textbf{Foot sliding} measures unnatural horizontal movement during ground contact. For each foot joint $j$ (ankles and toes) with height below threshold $H_j$, we compute:
\begin{equation}
    \text{Sliding}_j = {N_f} \sum_{t \in \mathcal{S}_j} \|\mathbf{p}_{j,t+1}^{xy} - \mathbf{p}_{j,t}^{xy}\|_2 \cdot (2 - 2^{(\mathbf{p}_{j,t}^{z}/H_j)})
\end{equation}
where $\mathbf{p}_{j,t}$ is the position of joint $j$ at frame $t$, $\mathcal{S}_j$ are frames where $\mathbf{p}_{j,t}^{z} < H_j$, and $N_f$ is the total frame count. The exponential weighting function gradually decreases influence as joints lift from the ground. The final metric averages across all four foot joints and is reported in centimeters. In a standard setting, the lower the foot sliding value, the more natural the motion.

\textbf{Jerk} quantifies motion smoothness by measuring the rate of change of acceleration. For a sequence of joint positions $\mathbf{p}$ with $N_f$ frames, we compute:
\begin{equation}
    \text{Jerk} = \frac{1}{N_f-3} \sum_{t=1}^{N_f-3} \|\mathbf{a}_{t+1} - \mathbf{a}_t\|_2,
\end{equation}
where velocities and accelerations are calculated as finite differences.
As indicated here, lower jerk represents more smooth motions.

\textbf{Motion Signal-to-Noise Ratio (MSNR)} quantifies motion quality through the SNR of joint kinematics, computed as:
\begin{equation}
    \text{SNR} = 10 \log_{10} \left( \frac{P_{\text{signal}}}{P_{\text{noise}}} \right) = 10 \log_{10} \left( \frac{\E[\hat{v}^2]}{\E[\abs{v-\hat{v}}^2]} \right),
\end{equation}
where $v$ represents the normalized local joint velocities, and $\hat{v}$ is the temporally smoothed version of $v$ obtained through convolution with a kernel size of 3. This metric captures the relationship between meaningful motion patterns and undesirable jitter or noise. A higher SNR value indicates a smoother motion. However, we should note that an overly smoothed signal may lose important details or contain less informative action.

\textbf{Coherence} score quantifies motion consistency by measuring pose cluster compactness. We compute coherence as 
\begin{equation}
    C = 1 - \frac{\mu_d}{\max_d},
\end{equation}
where $\mu_d$ is the mean distance from poses to their cluster centroids, and $\max_d$ is the maximum observed distance. Values approaching 1 indicate highly consistent movement patterns with minimal deviation.

\textbf{Diversity} metrics, on the other hand, quantify the variety of motion patterns in a dataset. We compute motion diversity using normalized Shannon entropy across pose clusters. After k-means clustering, we calculate
\begin{equation}
    D = -\frac{\sum_{i=1}^{n} p_i \log_2 p_i}{\log_2 n},
\end{equation}
where $p_i$ represents the proportion of frames in the $i$-th cluster. Higher diversity values indicate a wider range of motion patterns. However, this metric also identifies jittering or noise as diverse patterns.

\textbf{Penetration} quantifies the physical plausibility of human-object interactions by measuring object intrusion into the human mesh. For each frame, we sample points $\mathcal{P}_{obj}$ on object surfaces and compute the maximum penetration depth as:
\begin{equation}
    \text{Penetration}(t) = \min_{p \in \mathcal{P}_{obj}} d(p, \mathcal{M}_h),
\end{equation}
where $d(p, \mathcal{M}_h)$ is the signed distance from point $p$ to the human mesh $\mathcal{M}_h$. Positive distances indicate interior points, with more positive values representing deeper penetration. We report the average maximum penetration across all frames, with lower values indicating more physically plausible interactions.

\textbf{Contact entropy} quantifies the diversity of interaction states and transitions during human-object interaction. For a sequence of interaction states discretized into categories (large penetration, contact, proximity, and distance), we compute:
\begin{equation}
    \text{Entropy} = -\sum_{i, j} p(s_i \to s_j) \log_2 p(s_i \to s_j),
\end{equation}
where $p(s_i \to s_j)$ is the probability of transitioning from state $s_i$ to state $s_j$ across all sampled points and frames. Higher entropy values indicate more diverse and complex interactions, with a balanced distribution of different types of contact and approach behaviors.

\textbf{State consistency} measures the temporal stability of interaction states, rewarding smooth and persistent contacts while penalizing rapid state fluctuations. For each sampled point, we calculate the average run length normalized by sequence length:
\begin{equation}
    \text{Consistency} = \frac{1}{N_p} \sum_{p=1}^{N_p} \frac{\text{Avg. Run Length}_p}{\text{Sequence Length}}.
\end{equation}
We additionally penalize points with large penetrations by applying a scaling factor based on large penetration duration. Higher consistency scores indicate a more consistent interaction state with fewer state changes.

\subsection{Perceptual Evaluation Results}\label{sec:app_perceptual_eval}
\begin{figure*}[t]
    \centering
    \begin{subfigure}[t]{\textwidth}
        \centering
        \includegraphics[width=0.6\linewidth]{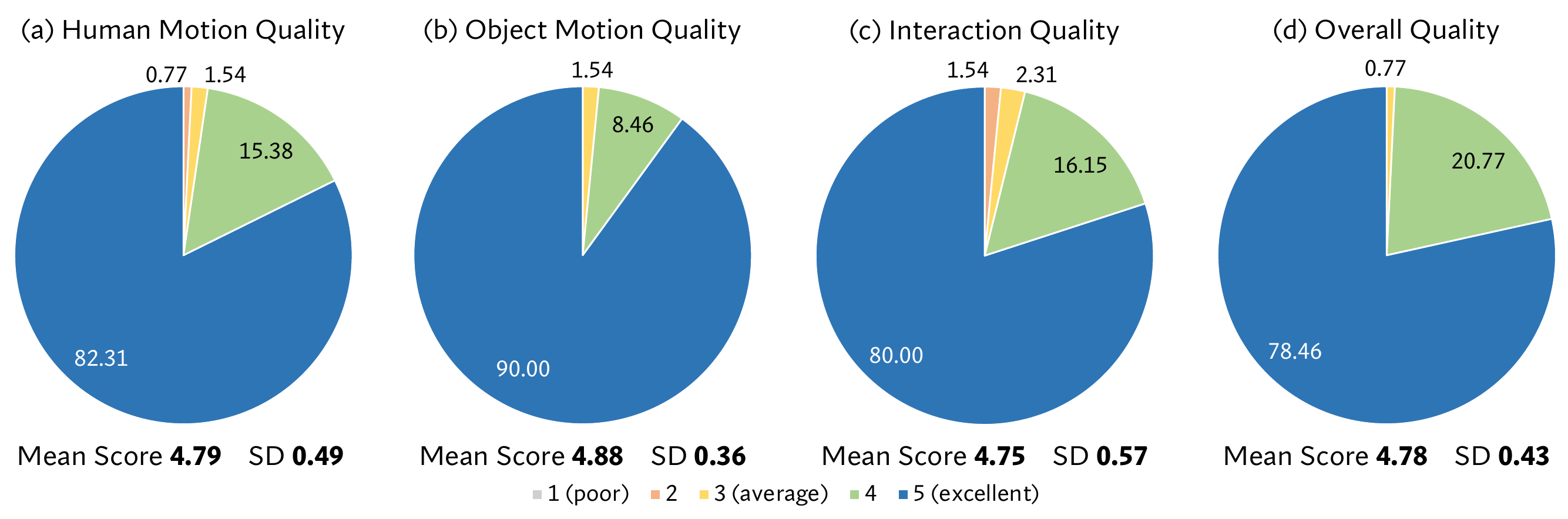}
        \caption{\name}
    \end{subfigure}%
    
    \begin{subfigure}[t]{\textwidth}
        \centering
        \includegraphics[width=0.6\linewidth]{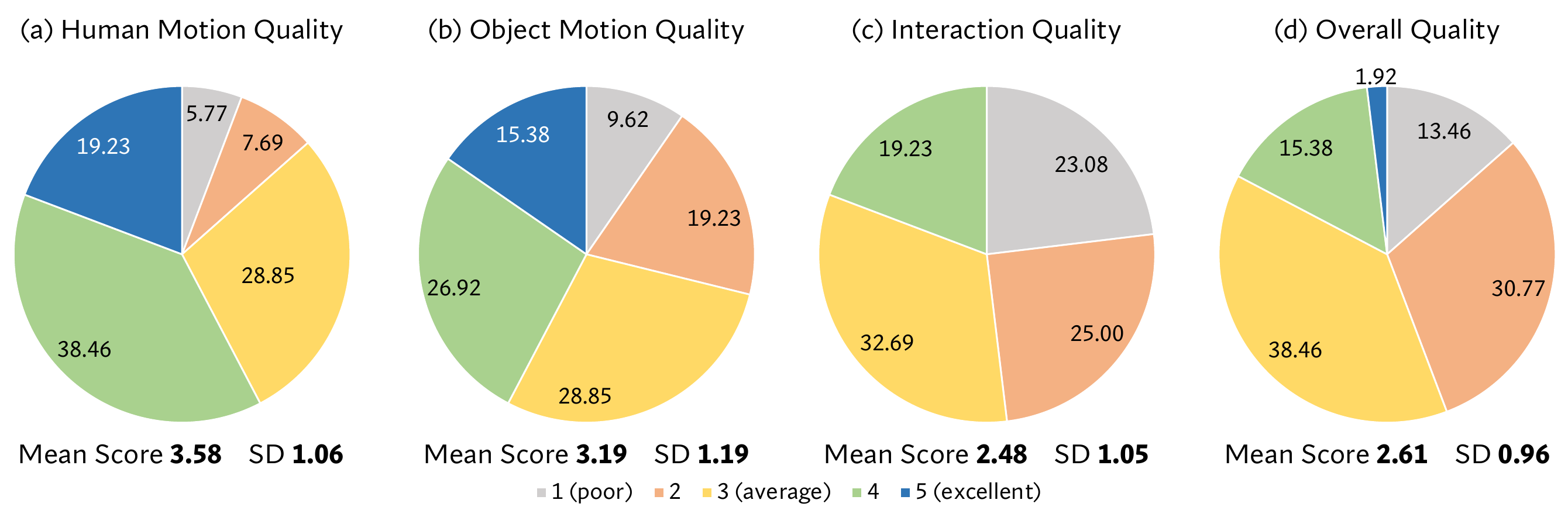}
        \caption{BEHAVE}
    \end{subfigure}%

    \begin{subfigure}[t]{\textwidth}
        \centering
        \includegraphics[width=0.6\linewidth]{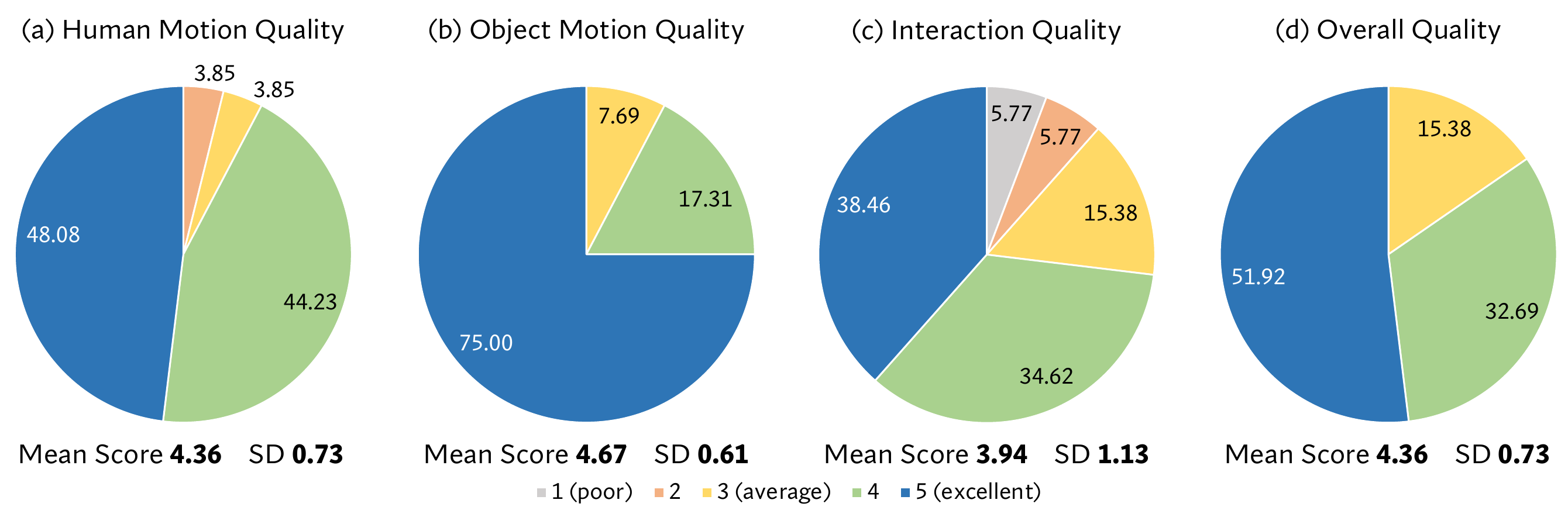}
        \caption{OMOMO}
    \end{subfigure}%

    \begin{subfigure}[t]{\textwidth}
        \centering
        \includegraphics[width=0.6\linewidth]{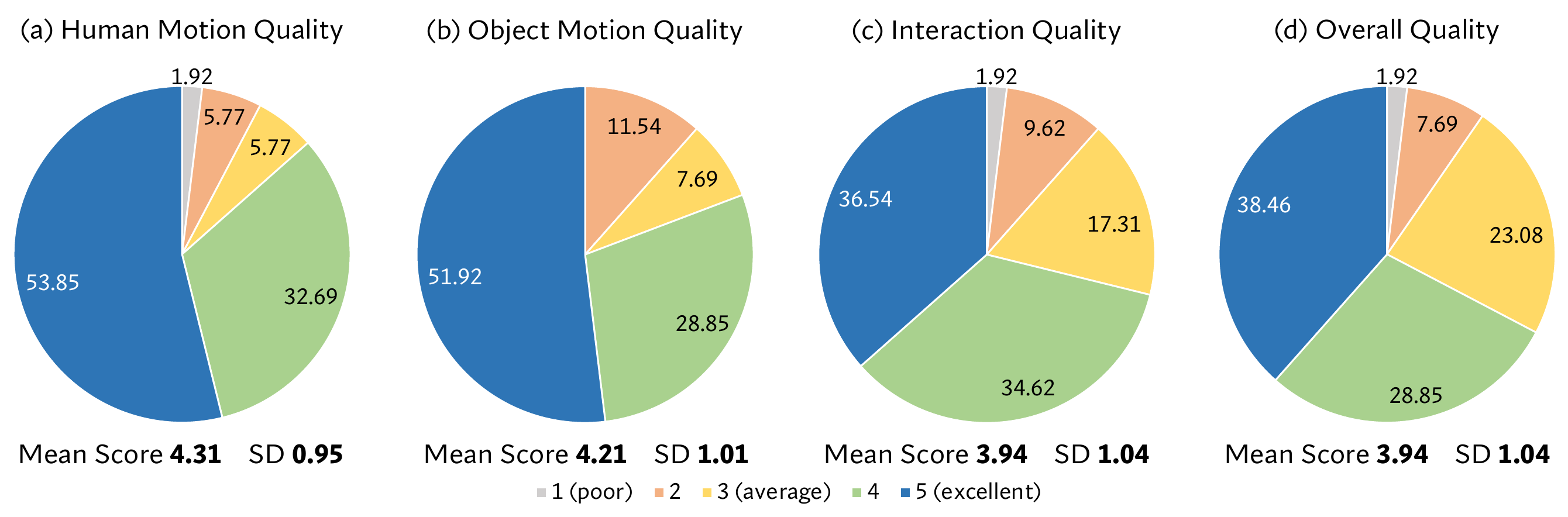}
        \caption{IMHD}
    \end{subfigure}%

    \begin{subfigure}[t]{\textwidth}
        \centering
        \includegraphics[width=0.6\linewidth]{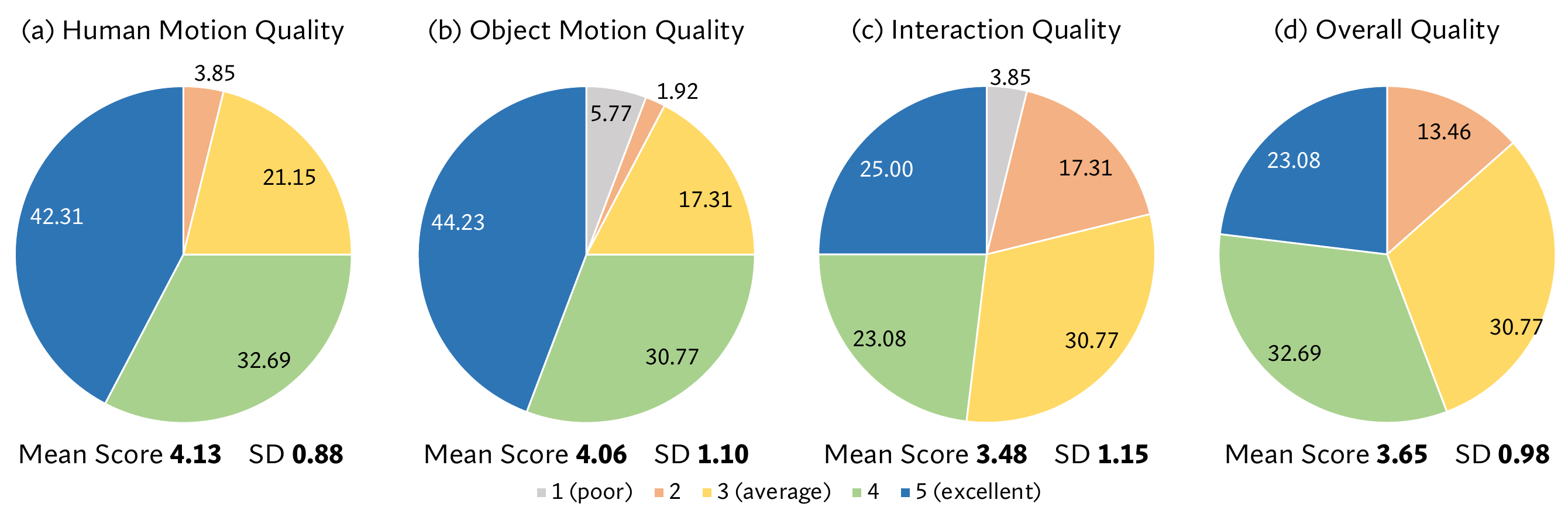}
        \caption{ParaHome}
    \end{subfigure}%
    \caption{\textbf{Perceptual absolute quality ratings.} We show the aggregate percentages of absolute quality ratings on five-point Likert scales from our participants for \name, BEHAVE~\cite{bhatnagar22behave}, OMOMO~\cite{Li2023omomo}, IMHD~\cite{zhao2024imhoi}, and ParaHome~\cite{Kim2024ParaHomePE}. We assess the quality on four aspects: \textit{(a) Human Motion Quality}, how plausible the human motions appear; \textit{(b) Object Motion Quality}, how plausible the object motions appear; \textit{(c) Interaction Quality}, how realistic the interactions between the humans and the objects appear; and \textit{(d) Overall Quality}, how realistic the overall animations appear. We observe significant increases in ratings of 5 for \name in all four aspects.}
    \label{fig:app_study_absolute}
\end{figure*}

\begin{figure*}[t]
    \centering
    \includegraphics[width=\linewidth]{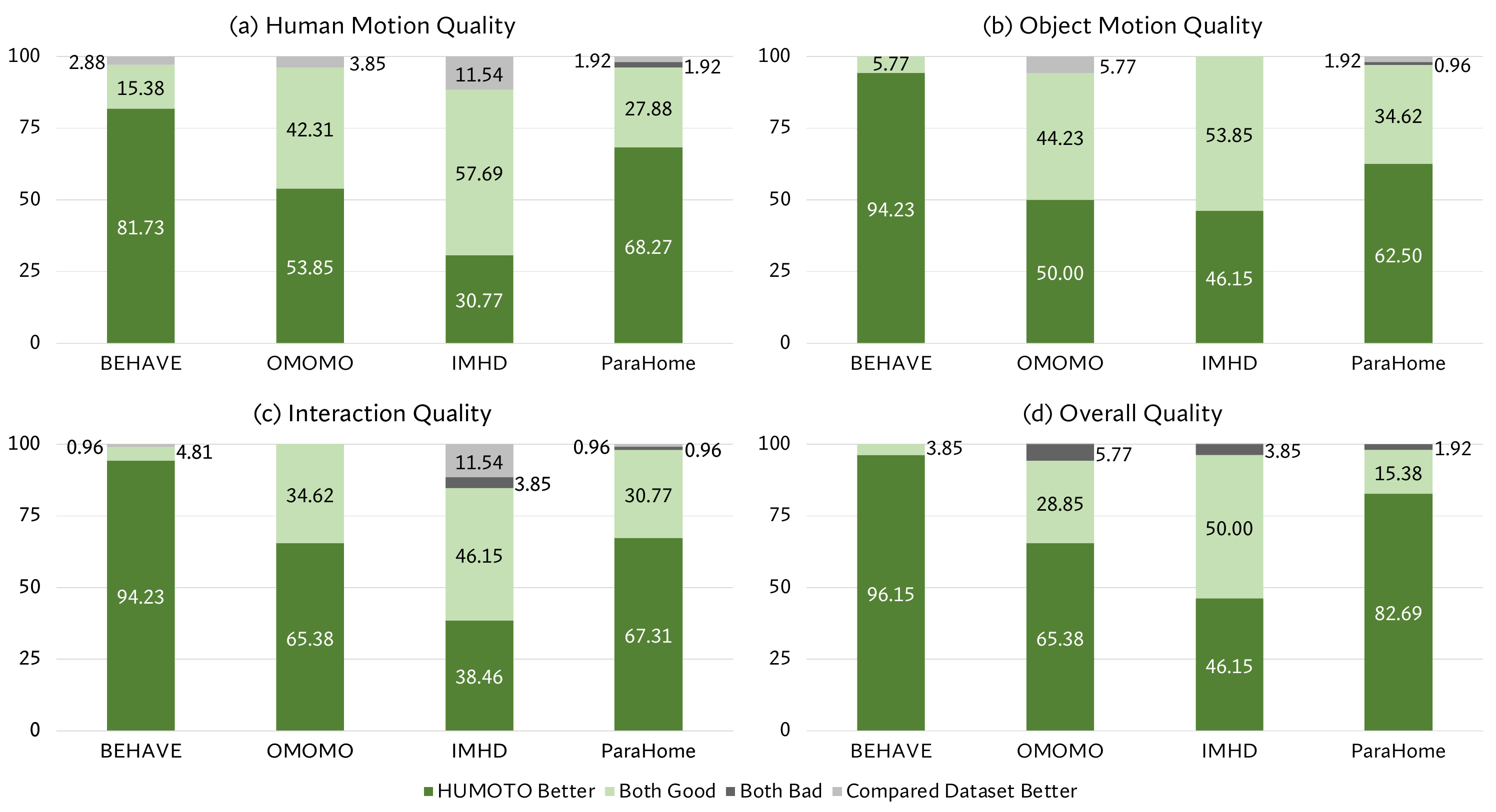}
    \caption{\textbf{Perceptual pairwise comparisons.} We show the aggregate percentages of pairwise comparison results from our participants, comparing side-by-side between \name and other datasets, including BEHAVE~\cite{bhatnagar22behave}, OMOMO~\cite{Li2023omomo}, IMHD~\cite{zhao2024imhoi}, and ParaHome~\cite{Kim2024ParaHomePE}. We assess the comparisons on four aspects: \textit{(a) Human Motion Quality}, how plausible the human motions appear; \textit{(b) Object Motion Quality}, how plausible the object motions appear; \textit{(c) Interaction Quality}, how realistic the interactions between the humans and the objects appear; and \textit{(d) Overall Quality}, how realistic the overall animations appear. After accounting for ties, we observe significant preferences for \name in all four aspects.}
    \label{fig:app_study_comparison}
\end{figure*}

Following the details of our perceptual study setup provided in \cref{sec:perceptual_study}, we provide the detailed score distribution percentages of absolute quality evaluations in \cref{fig:app_study_absolute} and pairwise evaluations in \cref{fig:app_study_comparison}.

\begin{figure*}[t]
    \centering
    \includegraphics[width=\linewidth]{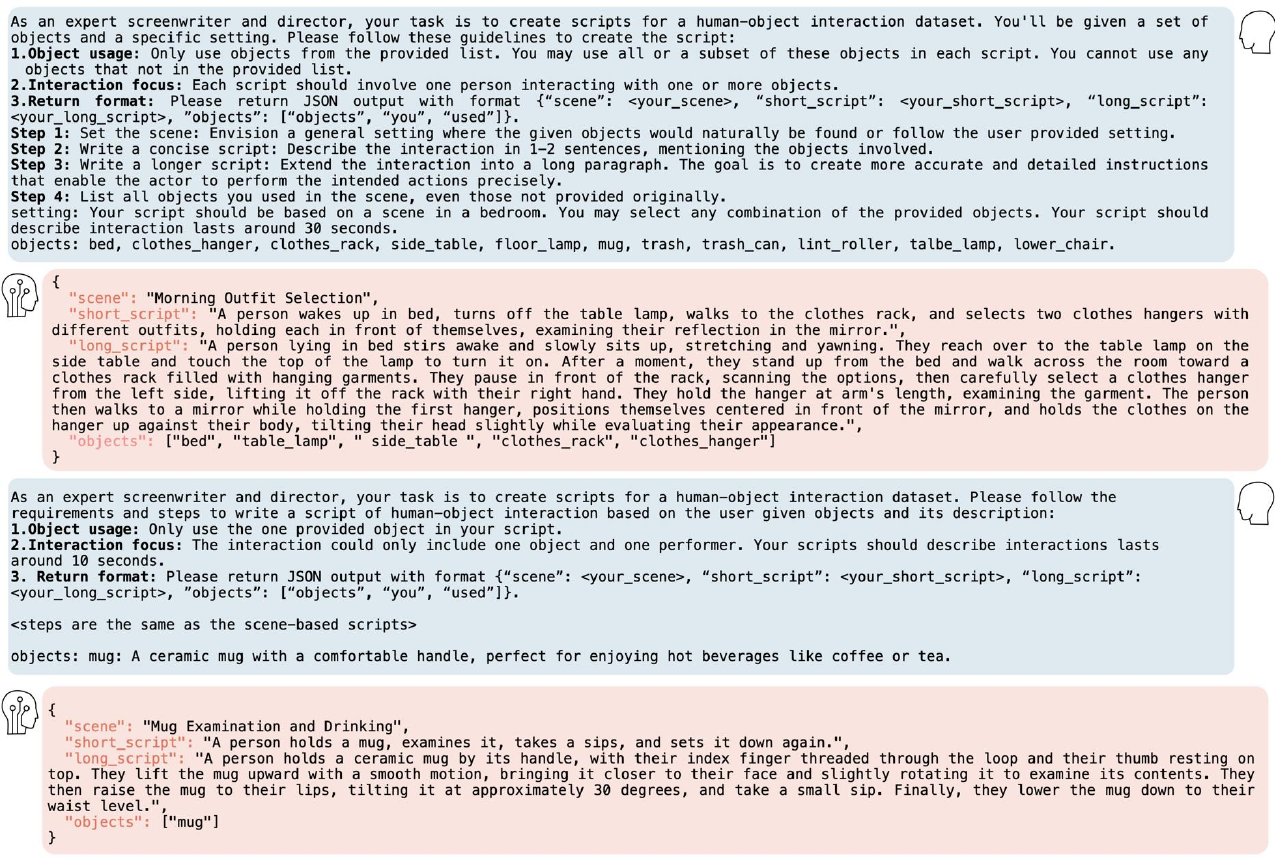}
    \caption{\textbf{Examples of how we use LLMs to develop our human-object interaction scripts for capturing.} \textit{Top:} We cluster objects into different scene types and create possible interactions within that scene. \textit{Bottom:} For each individual object, we prompt LLMs on how one person would be possible to interact with the object.}
    \vspace{-10pt}
    \label{fig:script_development}
    \vspace{-10pt}
\end{figure*}

\begin{figure*}[t]
    \centering
    \includegraphics[width=\linewidth]{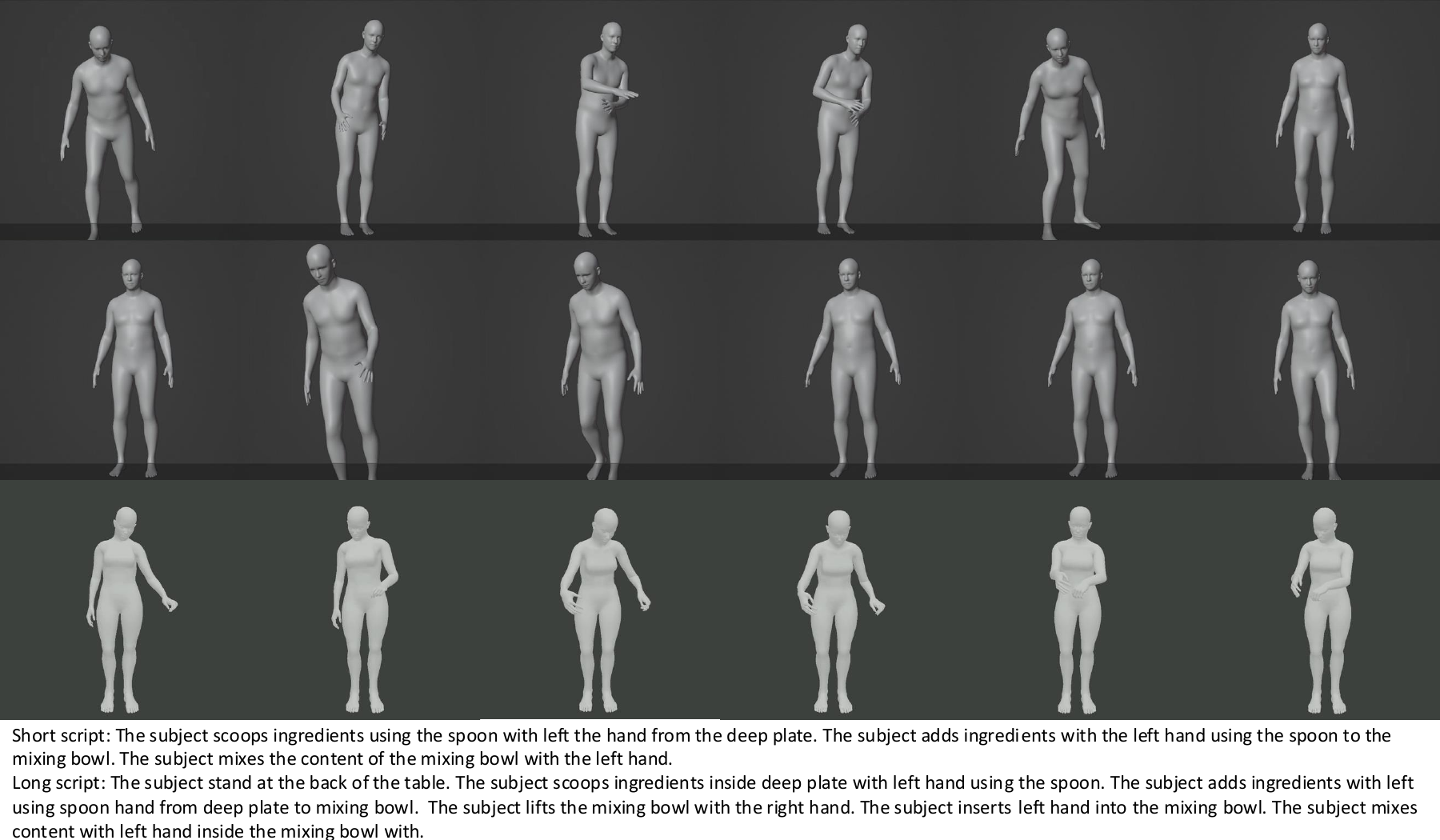}
    \caption{\textbf{Motion generation results comparing our text-annotated dataset with MotionGPT~\cite{jiang2024motiongpt}.} \textit{Top:} Generated motion sequence from short script input. \textit{Middle:} Generated motion sequence from detailed long script input. \textit{Bottom:} Ground truth motion sequence from our \name dataset. While MotionGPT can generate basic movements following general instructions, it struggles with the fine-grained hand-object interactions and precise manipulation sequences present in our dataset.}
    \label{fig:motiongpt_gen}
\end{figure*}


{
    \small
    \begin{thebibliography}{83}
\providecommand{\natexlab}[1]{#1}
\providecommand{\url}[1]{\texttt{#1}}
\expandafter\ifx\csname urlstyle\endcsname\relax
  \providecommand{\doi}[1]{doi: #1}\else
  \providecommand{\doi}{doi: \begingroup \urlstyle{rm}\Url}\fi

\bibitem[Bhatnagar et~al.(2022)Bhatnagar, Xie, Petrov, Sminchisescu, Theobalt, and Pons-Moll]{bhatnagar22behave}
Bharat~Lal Bhatnagar, Xianghui Xie, Ilya Petrov, Cristian Sminchisescu, Christian Theobalt, and Gerard Pons-Moll.
\newblock Behave: Dataset and method for tracking human object interactions.
\newblock In \emph{{IEEE} Conference on Computer Vision and Pattern Recognition (CVPR)}. {IEEE}, 2022.

\bibitem[Cao et~al.(2025)Cao, Pan, Han, Wong, and Liu]{cao2025avatargo}
Yukang Cao, Liang Pan, Kai Han, Kwan-Yee~K. Wong, and Ziwei Liu.
\newblock Avatar{GO}: Zero-shot 4d human-object interaction generation and animation.
\newblock In \emph{The Thirteenth International Conference on Learning Representations}, 2025.

\bibitem[Cao et~al.(2018)Cao, Hidalgo, Simon, Wei, and Sheikh]{Cao2018OpenPoseRM}
Zhe Cao, Gines Hidalgo, Tomas Simon, Shih-En Wei, and Yaser Sheikh.
\newblock Openpose: Realtime multi-person 2d pose estimation using part affinity fields.
\newblock \emph{IEEE Transactions on Pattern Analysis and Machine Intelligence}, 43:\penalty0 172--186, 2018.

\bibitem[Chi et~al.(2023)Chi, Xu, Feng, Cousineau, Du, Burchfiel, Tedrake, and Song]{chi2023diffusionpolicy}
Cheng Chi, Zhenjia Xu, Siyuan Feng, Eric Cousineau, Yilun Du, Benjamin Burchfiel, Russ Tedrake, and Shuran Song.
\newblock Diffusion policy: Visuomotor policy learning via action diffusion.
\newblock \emph{The International Journal of Robotics Research}, page 02783649241273668, 2023.

\bibitem[Corazza et~al.(2010)Corazza, M{\"u}ndermann, Gambaretto, Ferrigno, and Andriacchi]{15Corazza2010MarkerlessMC}
Stefano Corazza, Lars M{\"u}ndermann, Emiliano Gambaretto, Giancarlo Ferrigno, and Thomas~P. Andriacchi.
\newblock Markerless motion capture through visual hull, articulated icp and subject specific model generation.
\newblock \emph{International Journal of Computer Vision}, 87:\penalty0 156--169, 2010.

\bibitem[Coumans and Bai(2016)]{coumans2016pybullet}
Erwin Coumans and Yunfei Bai.
\newblock Pybullet, a python module for physics simulation for games, robotics and machine learning, 2016.

\bibitem[de~Aguiar et~al.(2008)de~Aguiar, Stoll, Theobalt, Ahmed, Seidel, and Thrun]{18performance}
Edilson de Aguiar, Carsten Stoll, Christian Theobalt, Naveed Ahmed, Hans-Peter Seidel, and Sebastian Thrun.
\newblock Performance capture from sparse multi-view video.
\newblock \emph{ACM Trans. Graph.}, 27\penalty0 (3):\penalty0 1–10, 2008.

\bibitem[DelPreto et~al.(2020)DelPreto, Lipton, Sanneman, Fay, Fourie, Choi, and Rus]{DelPreto2020HelpingRobotsLearn}
Joseph DelPreto, Jeffrey~Ian Lipton, Lindsay~M. Sanneman, Aidan~J. Fay, Christopher~K. Fourie, Changhyun Choi, and Daniela Rus.
\newblock Helping robots learn: A human-robot master-apprentice model using demonstrations via virtual reality teleoperation.
\newblock \emph{2020 IEEE International Conference on Robotics and Automation (ICRA)}, pages 10226--10233, 2020.

\bibitem[DelPreto et~al.(2022)DelPreto, Liu, Luo, Foshey, Li, Torralba, Matusik, and Rus]{19DelPreto2022ActionSenseAM}
Joseph DelPreto, Chao Liu, Yiyue Luo, Michael Foshey, Yunzhu Li, Antonio Torralba, Wojciech Matusik, and Daniela Rus.
\newblock Actionsense: A multimodal dataset and recording framework for human activities using wearable sensors in a kitchen environment.
\newblock In \emph{Neural Information Processing Systems}, 2022.

\bibitem[Diller and Dai(2024)]{diller2023cghoi}
Christian Diller and Angela Dai.
\newblock Cg-hoi: Contact-guided 3d human-object interaction generation.
\newblock In \emph{Proc. Computer Vision and Pattern Recognition (CVPR), IEEE}, 2024.

\bibitem[Fan et~al.(2023)Fan, Taheri, Tzionas, Kocabas, Kaufmann, Black, and Hilliges]{fan2023arctic}
Zicong Fan, Omid Taheri, Dimitrios Tzionas, Muhammed Kocabas, Manuel Kaufmann, Michael~J. Black, and Otmar Hilliges.
\newblock {ARCTIC}: A dataset for dexterous bimanual hand-object manipulation.
\newblock In \emph{Proceedings IEEE Conference on Computer Vision and Pattern Recognition (CVPR)}, 2023.

\bibitem[Fishman et~al.(2022)Fishman, Murali, Eppner, Peele, Boots, and Fox]{fishman2022mpinets}
Adam Fishman, Adithyavairavan Murali, Clemens Eppner, Bryan Peele, Byron Boots, and Dieter Fox.
\newblock Motion policy networks.
\newblock In \emph{Proceedings of the 6th Conference on Robot Learning (CoRL)}, 2022.

\bibitem[Fu et~al.(2024)Fu, Zhao, and Finn]{Fu2024MobileAloha}
Zipeng Fu, Tony Zhao, and Chelsea Finn.
\newblock Mobile aloha: Learning bimanual mobile manipulation with low-cost whole-body teleoperation.
\newblock \emph{ArXiv}, abs/2401.02117, 2024.

\bibitem[Furukawa and Ponce(2008)]{24Furukawa2008Dense3M}
Yasutaka Furukawa and Jean Ponce.
\newblock Dense 3d motion capture from synchronized video streams.
\newblock \emph{2008 IEEE Conference on Computer Vision and Pattern Recognition}, pages 1--8, 2008.

\bibitem[Gall et~al.(2009)Gall, Stoll, de~Aguiar, Theobalt, Rosenhahn, and Seidel]{25Gall2009MotionCU}
Juergen Gall, Carsten Stoll, Edilson de Aguiar, Christian Theobalt, Bodo Rosenhahn, and Hans-Peter Seidel.
\newblock Motion capture using joint skeleton tracking and surface estimation.
\newblock \emph{2009 IEEE Conference on Computer Vision and Pattern Recognition}, pages 1746--1753, 2009.

\bibitem[Gao et~al.(2024)Gao, Wang, Xiao, Wang, Wang, Cao, Hu, Liu, Dai, and Pang]{gao2024coohoi}
Jiawei Gao, Ziqin Wang, Zeqi Xiao, Jingbo Wang, Tai Wang, Jinkun Cao, Xiaolin Hu, Si Liu, Jifeng Dai, and Jiangmiao Pang.
\newblock Coo{HOI}: Learning cooperative human-object interaction with manipulated object dynamics.
\newblock In \emph{The Thirty-eighth Annual Conference on Neural Information Processing Systems}, 2024.

\bibitem[Garcia-Hernando et~al.(2017{\natexlab{a}})Garcia-Hernando, Yuan, Baek, and Kim]{26GarciaHernando2017FirstPersonHA}
Guillermo Garcia-Hernando, Shanxin Yuan, Seungryul Baek, and Tae-Kyun Kim.
\newblock First-person hand action benchmark with rgb-d videos and 3d hand pose annotations.
\newblock \emph{2018 IEEE/CVF Conference on Computer Vision and Pattern Recognition}, pages 409--419, 2017{\natexlab{a}}.

\bibitem[Garcia-Hernando et~al.(2017{\natexlab{b}})Garcia-Hernando, Yuan, Baek, and Kim]{GarciaHernando2017FirstPersonHA}
Guillermo Garcia-Hernando, Shanxin Yuan, Seungryul Baek, and Tae-Kyun Kim.
\newblock First-person hand action benchmark with rgb-d videos and 3d hand pose annotations.
\newblock \emph{2018 IEEE/CVF Conference on Computer Vision and Pattern Recognition}, pages 409--419, 2017{\natexlab{b}}.

\bibitem[Goel et~al.(2023)Goel, Pavlakos, Rajasegaran, Kanazawa, and Malik]{goel20234dhumans}
Shubham Goel, Georgios Pavlakos, Jathushan Rajasegaran, Angjoo Kanazawa, and Jitendra Malik.
\newblock Humans in 4{D}: Reconstructing and tracking humans with transformers.
\newblock In \emph{ICCV}, 2023.

\bibitem[G{\"u}ler et~al.(2018)G{\"u}ler, Neverova, and Kokkinos]{Gler2018DensePoseDH}
Riza~Alp G{\"u}ler, Natalia Neverova, and Iasonas Kokkinos.
\newblock Densepose: Dense human pose estimation in the wild.
\newblock \emph{2018 IEEE/CVF Conference on Computer Vision and Pattern Recognition}, pages 7297--7306, 2018.

\bibitem[Guo et~al.(2022)Guo, Zou, Zuo, Wang, Ji, Li, and Cheng]{Guo_2022_CVPR}
Chuan Guo, Shihao Zou, Xinxin Zuo, Sen Wang, Wei Ji, Xingyu Li, and Li Cheng.
\newblock Generating diverse and natural 3d human motions from text.
\newblock In \emph{Proceedings of the IEEE/CVF Conference on Computer Vision and Pattern Recognition (CVPR)}, pages 5152--5161, 2022.

\bibitem[Guzov et~al.(2022)Guzov, Chibane, Marin, He, Sattler, and Pons-Moll]{Guzov2022InteractionRT}
Vladimir Guzov, Julian Chibane, Riccardo Marin, Yannan He, Torsten Sattler, and Gerard Pons-Moll.
\newblock Interaction replica: Tracking human–object interaction and scene changes from human motion.
\newblock \emph{2024 International Conference on 3D Vision (3DV)}, pages 1006--1016, 2022.

\bibitem[Hampali et~al.(2019)Hampali, Rad, Oberweger, and Lepetit]{Hampali2019HOnnotateAM}
Shreyas Hampali, Mahdi Rad, Markus Oberweger, and Vincent Lepetit.
\newblock Honnotate: A method for 3d annotation of hand and object poses.
\newblock \emph{2020 IEEE/CVF Conference on Computer Vision and Pattern Recognition (CVPR)}, pages 3193--3203, 2019.

\bibitem[Han et~al.(2018)Han, Liu, Wang, Ye, Twigg, and Kin]{38Han2018OnlineOM}
Shangchen Han, Beibei Liu, Robert~Y. Wang, Yuting Ye, Christopher~D. Twigg, and Kenrick Kin.
\newblock Online optical marker-based hand tracking with deep labels.
\newblock \emph{ACM Transactions on Graphics (TOG)}, 37:\penalty0 1 -- 10, 2018.

\bibitem[Hassan et~al.(2023)Hassan, Guo, Wang, Black, Fidler, and Peng]{hassan2023synthphyint}
Mohamed Hassan, Yunrong Guo, Tingwu Wang, Michael Black, Sanja Fidler, and Xue~Bin Peng.
\newblock Synthesizing physical character-scene interactions.
\newblock In \emph{ACM SIGGRAPH 2023 Conference Proceedings}, New York, NY, USA, 2023. Association for Computing Machinery.

\bibitem[Hu(2024)]{Hu_2024_CVPR_animateanyone}
Li Hu.
\newblock Animate anyone: Consistent and controllable image-to-video synthesis for character animation.
\newblock In \emph{Proceedings of the IEEE/CVF Conference on Computer Vision and Pattern Recognition (CVPR)}, pages 8153--8163, 2024.

\bibitem[Hu et~al.(2024)Hu, Yin, Haeufle, Schmitt, and Bulling]{hu24hoimotion}
Zhiming Hu, Zheming Yin, Daniel Haeufle, Syn Schmitt, and Andreas Bulling.
\newblock Hoimotion: Forecasting human motion during human-object interactions using egocentric 3d object bounding boxes.
\newblock \emph{IEEE Transactions on Visualization and Computer Graphics}, 2024.

\bibitem[Huang et~al.(2022)Huang, Tehari, Black, for Intelligent~Systems, Tubingen, Germany, of~Amsterdam, Amsterdam, and Netherlands.]{Huang2022InterCapJM}
Yinghao Huang, Omid Tehari, Michael~J. Black, Dimitrios Tzionas Max Planck~Institute for Intelligent~Systems, Tubingen, Germany, University of Amsterdam, Amsterdam, and The Netherlands.
\newblock Intercap: Joint markerless 3d tracking of humans and objects in interaction.
\newblock In \emph{German Conference on Pattern Recognition}, 2022.

\bibitem[Inc.(2018)]{mixamo}
Adobe~Systems Inc.
\newblock Mixamo, 2018.
\newblock Accessed: 2025-03-07.

\bibitem[Jiang et~al.(2024{\natexlab{a}})Jiang, Chen, Liu, Yu, Yu, and Chen]{jiang2024motiongpt}
Biao Jiang, Xin Chen, Wen Liu, Jingyi Yu, Gang Yu, and Tao Chen.
\newblock Motiongpt: Human motion as a foreign language.
\newblock \emph{Advances in Neural Information Processing Systems}, 36, 2024{\natexlab{a}}.

\bibitem[Jiang et~al.(2022)Jiang, Liu, Cao, Cui, Zhang, Chen, Wang, Zhu, and Huang]{Jiang2022FullBodyAH}
Nan Jiang, Tengyu Liu, Zhexuan Cao, Jieming Cui, Zhiyuan Zhang, Yixin Chen, Heng Wang, Yixin Zhu, and Siyuan Huang.
\newblock Full-body articulated human-object interaction.
\newblock \emph{2023 IEEE/CVF International Conference on Computer Vision (ICCV)}, pages 9331--9342, 2022.

\bibitem[Jiang et~al.(2024{\natexlab{b}})Jiang, He, Wang, Li, Chen, Huang, and Zhu]{jiang2024autocharacterscene}
Nan Jiang, Zimo He, Zi Wang, Hongjie Li, Yixin Chen, Siyuan Huang, and Yixin Zhu.
\newblock Autonomous character-scene interaction synthesis from text instruction.
\newblock In \emph{SIGGRAPH Asia 2024 Conference Papers}, New York, NY, USA, 2024{\natexlab{b}}. Association for Computing Machinery.

\bibitem[Jiang et~al.(2024{\natexlab{c}})Jiang, Zhang, Li, Ma, Wang, Chen, Liu, Zhu, and Huang]{jiang2024TRUMANS}
Nan Jiang, Zhiyuan Zhang, Hongjie Li, Xiaoxuan Ma, Zan Wang, Yixin Chen, Tengyu Liu, Yixin Zhu, and Siyuan Huang.
\newblock Scaling up dynamic human-scene interaction modeling.
\newblock In \emph{Proceedings of the IEEE/CVF Conference on Computer Vision and Pattern Recognition}, pages 1737--1747, 2024{\natexlab{c}}.

\bibitem[Kehl and Gool(2006)]{56Kehl2006MarkerlessTO}
Roland Kehl and Luc~Van Gool.
\newblock Markerless tracking of complex human motions from multiple views.
\newblock \emph{Comput. Vis. Image Underst.}, 104:\penalty0 190--209, 2006.

\bibitem[Kim et~al.(2024)Kim, Kim, Na, and Joo]{Kim2024ParaHomePE}
Jeonghwan Kim, Jisoo Kim, Jeonghyeon Na, and Hanbyul Joo.
\newblock Parahome: Parameterizing everyday home activities towards 3d generative modeling of human-object interactions.
\newblock \emph{ArXiv}, abs/2401.10232, 2024.

\bibitem[Kim et~al.(2016)Kim, Park, Bang, and Lee]{Kim2016RetargetingHoi}
Yeonjoon Kim, Hangi Park, Seungbae Bang, and Sung-Hee Lee.
\newblock Retargeting human-object interaction to virtual avatars.
\newblock \emph{IEEE Transactions on Visualization and Computer Graphics}, 22:\penalty0 2405--2412, 2016.

\bibitem[Kobayashi et~al.(2023)Kobayashi, Liao, Inoue, Yojima, and Takahashi]{kobayashi2023motion}
Makito Kobayashi, Chen-Chieh Liao, Keito Inoue, Sentaro Yojima, and Masafumi Takahashi.
\newblock Motion capture dataset for practical use of ai-based motion editing and stylization, 2023.

\bibitem[Kovar et~al.(2002)Kovar, Gleicher, and Pighin]{Kovar2002MotionGraph}
Lucas Kovar, Michael Gleicher, and Fr{\'e}d{\'e}ric~H. Pighin.
\newblock Motion graphs.
\newblock \emph{Seminal Graphics Papers: Pushing the Boundaries, Volume 2}, 2002.

\bibitem[Li et~al.(2023)Li, Wu, and Liu]{Li2023omomo}
Jiaman Li, Jiajun Wu, and C.~Karen Liu.
\newblock Object motion guided human motion synthesis.
\newblock \emph{ACM Transactions on Graphics (TOG)}, 42:\penalty0 1 -- 11, 2023.

\bibitem[Li et~al.(2024{\natexlab{a}})Li, Clegg, Mottaghi, Wu, Puig, and Liu]{li2023controllablehoisynth}
Jiaman Li, Alexander Clegg, Roozbeh Mottaghi, Jiajun Wu, Xavier Puig, and C.~Karen Liu.
\newblock Controllable human-object interaction synthesis.
\newblock In \emph{ECCV}, 2024{\natexlab{a}}.

\bibitem[Li et~al.(2024{\natexlab{b}})Li, Wang, and Yang]{li2024humanobject}
Liulei Li, Wenguan Wang, and Yi Yang.
\newblock Human-object interaction detection collaborated with large relation-driven diffusion models.
\newblock In \emph{The Thirty-eighth Annual Conference on Neural Information Processing Systems}, 2024{\natexlab{b}}.

\bibitem[Lin et~al.(2023)Lin, Liu, Lu, and Jia]{lin2023sam6d}
Jiehong Lin, Lihua Liu, Dekun Lu, and Kui Jia.
\newblock Sam-6d: Segment anything model meets zero-shot 6d object pose estimation.
\newblock \emph{arXiv preprint arXiv:2311.15707}, 2023.

\bibitem[Liu et~al.(2022)Liu, Liu, Jiang, Fu, Lyu, Wan, Shen, Liang, Wang, and Yi]{Liu2022HOI4DA4}
Yunze Liu, Yun Liu, Chen Jiang, Zhoujie Fu, Kangbo Lyu, Weikang Wan, Hao Shen, Bo-Hua Liang, He Wang, and Li Yi.
\newblock Hoi4d: A 4d egocentric dataset for category-level human-object interaction.
\newblock \emph{2022 IEEE/CVF Conference on Computer Vision and Pattern Recognition (CVPR)}, pages 20981--20990, 2022.

\bibitem[Liu et~al.(2024)Liu, Yang, Si, Liu, Li, Zhang, Liu, and Yi]{liu2024taco}
Yun Liu, Haolin Yang, Xu Si, Ling Liu, Zipeng Li, Yuxiang Zhang, Yebin Liu, and Li Yi.
\newblock Taco: Benchmarking generalizable bimanual tool-action-object understanding.
\newblock \emph{arXiv preprint arXiv:2401.08399}, 2024.

\bibitem[Loper et~al.(2014)Loper, Mahmood, and Black]{Loper:SIGASIA:2014}
Matthew~M. Loper, Naureen Mahmood, and Michael~J. Black.
\newblock {MoSh}: Motion and shape capture from sparse markers.
\newblock \emph{ACM Transactions on Graphics, (Proc. SIGGRAPH Asia)}, 33\penalty0 (6):\penalty0 220:1--220:13, 2014.

\bibitem[Lu et~al.(2024)Lu, Kang, Li, Liu, Yang, Huang, and Hua]{lu2024ugg}
Jiaxin Lu, Hao Kang, Haoxiang Li, Bo Liu, Yiding Yang, Qixing Huang, and Gang Hua.
\newblock Ugg: Unified generative grasping.
\newblock In \emph{Computer Vision – ECCV 2024: 18th European Conference, Milan, Italy, September 29–October 4, 2024, Proceedings, Part LXVII}, page 414–433, Berlin, Heidelberg, 2024. Springer-Verlag.

\bibitem[Mahmood et~al.(2019)Mahmood, Ghorbani, Troje, Pons-Moll, and Black]{AMASS:2019}
Naureen Mahmood, Nima Ghorbani, Nikolaus~F. Troje, Gerard Pons-Moll, and Michael~J. Black.
\newblock Amass: Archive of motion capture as surface shapes.
\newblock In \emph{The IEEE International Conference on Computer Vision (ICCV)}, 2019.

\bibitem[Mascaro et~al.(2023)Mascaro, Sliwowski, and Lee]{mascaro2023hoiabot}
Esteve~Valls Mascaro, Daniel Sliwowski, and Dongheui Lee.
\newblock {HOI}4{ABOT}: Human-object interaction anticipation for human intention reading collaborative ro{BOT}s.
\newblock In \emph{7th Annual Conference on Robot Learning}, 2023.

\bibitem[Nam et~al.(2024)Nam, Jung, Moon, and Lee]{Nam_2024_CVPR}
Hyeongjin Nam, Daniel~Sungho Jung, Gyeongsik Moon, and Kyoung~Mu Lee.
\newblock Joint reconstruction of 3d human and object via contact-based refinement transformer.
\newblock In \emph{Proceedings of the IEEE/CVF Conference on Computer Vision and Pattern Recognition (CVPR)}, pages 10218--10227, 2024.

\bibitem[Park et~al.(2023)Park, Park, and Lee]{Park_2023_CVPR}
Jeeseung Park, Jin-Woo Park, and Jong-Seok Lee.
\newblock Viplo: Vision transformer based pose-conditioned self-loop graph for human-object interaction detection.
\newblock In \emph{Proceedings of the IEEE/CVF Conference on Computer Vision and Pattern Recognition (CVPR)}, pages 17152--17162, 2023.

\bibitem[Peng et~al.(2023)Peng, Xie, Wu, Jampani, Sun, and Jiang]{peng2023hoidiff}
Xiaogang Peng, Yiming Xie, Zizhao Wu, Varun Jampani, Deqing Sun, and Huaizu Jiang.
\newblock Hoi-diff: Text-driven synthesis of 3d human-object interactions using diffusion models.
\newblock \emph{arXiv preprint arXiv:2312.06553}, 2023.

\bibitem[Peng et~al.(2018)Peng, Abbeel, Levine, and van~de Panne]{2018-TOG-deepMimic}
Xue~Bin Peng, Pieter Abbeel, Sergey Levine, and Michiel van~de Panne.
\newblock Deepmimic: Example-guided deep reinforcement learning of physics-based character skills.
\newblock \emph{ACM Trans. Graph.}, 37\penalty0 (4):\penalty0 143:1--143:14, 2018.

\bibitem[{Rokoko}(2022)]{rokoko_smartsuit_pro_ii}
{Rokoko}.
\newblock Smartsuit pro ii.
\newblock \url{https://www.rokoko.com/products/smartsuit-pro}, 2022.

\bibitem[Sener et~al.(2022)Sener, Chatterjee, Shelepov, He, Singhania, Wang, and Yao]{Sener2022Assembly101AL}
Fadime Sener, Dibyadip Chatterjee, Daniel Shelepov, Kun He, Dipika Singhania, Robert Wang, and Angela Yao.
\newblock Assembly101: A large-scale multi-view video dataset for understanding procedural activities.
\newblock \emph{2022 IEEE/CVF Conference on Computer Vision and Pattern Recognition (CVPR)}, pages 21064--21074, 2022.

\bibitem[Shannon(1949)]{Shannon1949CommunicationIT}
Claude~E. Shannon.
\newblock Communication in the presence of noise.
\newblock \emph{Proceedings of the IRE}, 37:\penalty0 10--21, 1949.

\bibitem[Starke et~al.(2019)Starke, Zhang, Komura, and Saito]{starke2019nsm}
Sebastian Starke, He Zhang, Taku Komura, and Jun Saito.
\newblock Neural state machine for character-scene interactions.
\newblock \emph{ACM Trans. Graph.}, 38\penalty0 (6), 2019.

\bibitem[Stoll et~al.(2011)Stoll, Hasler, Gall, Seidel, and Theobalt]{90Stoll2011FastAM}
Carsten Stoll, Nils Hasler, Juergen Gall, Hans-Peter Seidel, and Christian Theobalt.
\newblock Fast articulated motion tracking using a sums of gaussians body model.
\newblock \emph{2011 International Conference on Computer Vision}, pages 951--958, 2011.

\bibitem[Taheri et~al.(2020)Taheri, Ghorbani, Black, and Tzionas]{GRAB:2020}
Omid Taheri, Nima Ghorbani, Michael~J. Black, and Dimitrios Tzionas.
\newblock {GRAB}: A dataset of whole-body human grasping of objects.
\newblock In \emph{European Conference on Computer Vision (ECCV)}, 2020.

\bibitem[Tian et~al.(2022)Tian, Zhao, Liu, Liu, Mao, Zhao, and Yan]{Tian2022SAMPAM}
Rong Tian, Zijing Zhao, Weijie Liu, Haoyan Liu, Weiquan Mao, Zhe Zhao, and Kimmo Yan.
\newblock Samp: A model inference toolkit of post-training quantization for text processing via self-adaptive mixed-precision.
\newblock In \emph{Conference on Empirical Methods in Natural Language Processing}, 2022.

\bibitem[Tremblay et~al.(2018)Tremblay, To, Sundaralingam, Xiang, Fox, and Birchfield]{Tremblay2018DeepOP}
Jonathan Tremblay, Thang To, Balakumar Sundaralingam, Yu Xiang, Dieter Fox, and Stan Birchfield.
\newblock Deep object pose estimation for semantic robotic grasping of household objects.
\newblock \emph{ArXiv}, abs/1809.10790, 2018.

\bibitem[Wan et~al.(2022)Wan, Yang, Liu, Zhang, Jia, Choi, Pan, Theobalt, Komura, and Wang]{wan2022learn}
Weilin Wan, Lei Yang, Lingjie Liu, Zhuoying Zhang, Ruixing Jia, Yi-King Choi, Jia Pan, Christian Theobalt, Taku Komura, and Wenping Wang.
\newblock Learn to predict how humans manipulate large-sized objects from interactive motions.
\newblock \emph{IEEE Robotics and Automation Letters}, 7\penalty0 (2):\penalty0 4702--4709, 2022.

\bibitem[Wan et~al.(2023)Wan, Geng, Liu, Shan, Yang, Yi, and Wang]{Wan_2023_unidexgrasppp}
Weikang Wan, Haoran Geng, Yun Liu, Zikang Shan, Yaodong Yang, Li Yi, and He Wang.
\newblock Unidexgrasp++: Improving dexterous grasping policy learning via geometry-aware curriculum and iterative generalist-specialist learning.
\newblock In \emph{Proceedings of the IEEE/CVF International Conference on Computer Vision (ICCV)}, pages 3891--3902, 2023.

\bibitem[Wang et~al.(2022)Wang, Zhang, Chen, Xu, Li, Liu, and Wang]{Wang2022DexGraspNetAL}
Ruicheng Wang, Jialiang Zhang, Jiayi Chen, Yinzhen Xu, Puhao Li, Tengyu Liu, and He Wang.
\newblock Dexgraspnet: A large-scale robotic dexterous grasp dataset for general objects based on simulation.
\newblock \emph{2023 IEEE International Conference on Robotics and Automation (ICRA)}, pages 11359--11366, 2022.

\bibitem[Wang et~al.(2024)Wang, Wang, Liu, and Daniilidis]{Wang2024TRAMGT}
Yufu Wang, Ziyun Wang, Lingjie Liu, and Kostas Daniilidis.
\newblock Tram: Global trajectory and motion of 3d humans from in-the-wild videos.
\newblock In \emph{European Conference on Computer Vision}, 2024.

\bibitem[Wei et~al.(2022)Wei, Liu, Ling, and Su]{wei2022coacd}
Xinyue Wei, Minghua Liu, Zhan Ling, and Hao Su.
\newblock Approximate convex decomposition for 3d meshes with collision-aware concavity and tree search.
\newblock \emph{ACM Transactions on Graphics (TOG)}, 41\penalty0 (4):\penalty0 1--18, 2022.

\bibitem[Wen et~al.(2024)Wen, Yang, Kautz, and Birchfield]{Wen_2024_CVPR_foundationpose}
Bowen Wen, Wei Yang, Jan Kautz, and Stan Birchfield.
\newblock Foundationpose: Unified 6d pose estimation and tracking of novel objects.
\newblock In \emph{Proceedings of the IEEE/CVF Conference on Computer Vision and Pattern Recognition (CVPR)}, pages 17868--17879, 2024.

\bibitem[Wu et~al.(2024)Wu, Li, Xu, and Liu]{wu2024humanobjectinteractionhumanlevelinstructions}
Zhen Wu, Jiaman Li, Pei Xu, and C.~Karen Liu.
\newblock Human-object interaction from human-level instructions, 2024.

\bibitem[Xiang et~al.(2017)Xiang, Schmidt, Narayanan, and Fox]{Xiang2017PoseCNNAC}
Yu Xiang, Tanner Schmidt, Venkatraman Narayanan, and Dieter Fox.
\newblock Posecnn: A convolutional neural network for 6d object pose estimation in cluttered scenes.
\newblock \emph{ArXiv}, abs/1711.00199, 2017.

\bibitem[Xie et~al.(2022)Xie, Bhatnagar, and Pons-Moll]{xie2022chore}
Xianghui Xie, Bharat~Lal Bhatnagar, and Gerard Pons-Moll.
\newblock Chore: Contact, human and object reconstruction from a single rgb image.
\newblock In \emph{European Conference on Computer Vision ({ECCV})}. {Springer}, 2022.

\bibitem[Xie et~al.(2024)Xie, Wang, Athanasiou, Bhatnagar, Huang, Mo, Chen, Jia, Zhang, Cui, Lin, Qian, Xiao, Yang, Nam, Jung, Kim, Lee, Hilliges, and Pons-Moll]{xie2024rhobinchallengereconstructionhuman}
Xianghui Xie, Xi Wang, Nikos Athanasiou, Bharat~Lal Bhatnagar, Chun-Hao~P. Huang, Kaichun Mo, Hao Chen, Xia Jia, Zerui Zhang, Liangxian Cui, Xiao Lin, Bingqiao Qian, Jie Xiao, Wenfei Yang, Hyeongjin Nam, Daniel~Sungho Jung, Kihoon Kim, Kyoung~Mu Lee, Otmar Hilliges, and Gerard Pons-Moll.
\newblock Rhobin challenge: Reconstruction of human object interaction, 2024.

\bibitem[Xu et~al.(2023)Xu, Li, Wang, and Gui]{xu2023interdiff}
Sirui Xu, Zhengyuan Li, Yu-Xiong Wang, and Liang-Yan Gui.
\newblock {InterDiff}: Generating 3d human-object interactions with physics-informed diffusion.
\newblock In \emph{ICCV}, 2023.

\bibitem[Xu et~al.(2024)Xu, Wang, Wang, and Gui]{xu2024interdreamer}
Sirui Xu, Ziyin Wang, Yu-Xiong Wang, and Liangyan Gui.
\newblock Interdreamer: Zero-shot text to 3d dynamic human-object interaction.
\newblock In \emph{The Thirty-eighth Annual Conference on Neural Information Processing Systems}, 2024.

\bibitem[Xue et~al.(2024)Xue, Luo, Chen, and Grauman]{xue2024hoiswap}
Zihui Xue, Mi Luo, Changan Chen, and Kristen Grauman.
\newblock {HOI}-swap: Swapping objects in videos with hand-object interaction awareness.
\newblock In \emph{The Thirty-eighth Annual Conference on Neural Information Processing Systems}, 2024.

\bibitem[Ye et~al.(2023)Ye, Li, Gupta, Mello, Birchfield, Song, Tulsiani, and Liu]{ye2023affordance}
Yufei Ye, Xueting Li, Abhinav Gupta, Shalini~De Mello, Stan Birchfield, Jiaming Song, Shubham Tulsiani, and Sifei Liu.
\newblock Affordance diffusion: Synthesizing hand-object interactions.
\newblock In \emph{CVPR}, 2023.

\bibitem[Yuan et~al.(2023)Yuan, Zhang, Wang, Albanie, Pan, Feng, Jiang, Ni, Zhang, and Zhao]{Yuan_2023_ICCV}
Hangjie Yuan, Shiwei Zhang, Xiang Wang, Samuel Albanie, Yining Pan, Tao Feng, Jianwen Jiang, Dong Ni, Yingya Zhang, and Deli Zhao.
\newblock Rlipv2: Fast scaling of relational language-image pre-training.
\newblock In \emph{Proceedings of the IEEE/CVF International Conference on Computer Vision (ICCV)}, pages 21649--21661, 2023.

\bibitem[Zhang et~al.(2023{\natexlab{a}})Zhang, Yuan, Campbell, Zhong, and Gould]{Zhang_2023_ICCV}
Frederic~Z Zhang, Yuhui Yuan, Dylan Campbell, Zhuoyao Zhong, and Stephen Gould.
\newblock Exploring predicate visual context in detecting of human-object interactions.
\newblock In \emph{Proceedings of the IEEE/CVF International Conference on Computer Vision (ICCV)}, pages 10411--10421, 2023{\natexlab{a}}.

\bibitem[Zhang et~al.(2023{\natexlab{b}})Zhang, Luo, Yang, Xu, Wu, Shi, Yu, Xu, and Wang]{zhang2023neuraldome}
Juze Zhang, Haimin Luo, Hongdi Yang, Xinru Xu, Qianyang Wu, Ye Shi, Jingyi Yu, Lan Xu, and Jingya Wang.
\newblock Neuraldome: A neural modeling pipeline on multi-view human-object interactions.
\newblock In \emph{CVPR}, 2023{\natexlab{b}}.

\bibitem[Zhang et~al.(2024)Zhang, Zhang, Song, Shi, Zhao, Shi, Yu, Xu, and Wang]{Zhang2024HOIM3CM}
Juze Zhang, Jingyan Zhang, Zining Song, Zhanhe Shi, Chengfeng Zhao, Ye Shi, Jingyi Yu, Lan Xu, and Jingya Wang.
\newblock Hoi-m3: Capture multiple humans and objects interaction within contextual environment.
\newblock \emph{2024 IEEE/CVF Conference on Computer Vision and Pattern Recognition (CVPR)}, pages 516--526, 2024.

\bibitem[Zhang et~al.(2020)Zhang, Pepose, Joo, Ramanan, Malik, and Kanazawa]{zhang2020phosa}
Jason~Y. Zhang, Sam Pepose, Hanbyul Joo, Deva Ramanan, Jitendra Malik, and Angjoo Kanazawa.
\newblock Perceiving 3d human-object spatial arrangements from a single image in the wild.
\newblock In \emph{European Conference on Computer Vision (ECCV)}, 2020.

\bibitem[Zhang et~al.(2022{\natexlab{a}})Zhang, Bhatnagar, Guzov, Starke, and Pons-Moll]{Zhang2022COUCHTC}
Xiaohan Zhang, Bharat~Lal Bhatnagar, Vladimir Guzov, Sebastian Starke, and Gerard Pons-Moll.
\newblock Couch: Towards controllable human-chair interactions.
\newblock \emph{ArXiv}, abs/2205.00541, 2022{\natexlab{a}}.

\bibitem[Zhang et~al.(2022{\natexlab{b}})Zhang, Bhatnagar, Starke, Guzov, and Pons-Moll]{zhang2022couch}
Xiaohan Zhang, Bharat~Lal Bhatnagar, Sebastian Starke, Vladimir Guzov, and Gerard Pons-Moll.
\newblock Couch: Towards controllable human-chair interactions.
\newblock In \emph{European Conference on Computer Vision ({ECCV})}. {Springer}, 2022{\natexlab{b}}.

\bibitem[Zhao et~al.(2024)Zhao, Zhang, Du, Shan, Wang, Yu, Wang, and Xu]{zhao2024imhoi}
Chengfeng Zhao, Juze Zhang, Jiashen Du, Ziwei Shan, Junye Wang, Jingyi Yu, Jingya Wang, and Lan Xu.
\newblock I'm hoi: Inertia-aware monocular capture of 3d human-object interactions.
\newblock In \emph{Proceedings of the IEEE/CVF Conference on Computer Vision and Pattern Recognition (CVPR)}, pages 729--741, 2024.

\bibitem[Zhu et~al.(2023)Zhu, Samet, and Picard]{human36m}
Yue Zhu, Nermin Samet, and David Picard.
\newblock H3wb: Human3.6m 3d wholebody dataset and benchmark.
\newblock In \emph{Proceedings of the IEEE/CVF International Conference on Computer Vision (ICCV)}, pages 20166--20177, 2023.

\end{thebibliography}

}
\end{document}